%% file: main.tex
\pdfoutput=1
\documentclass{article}

\PassOptionsToPackage{numbers,sort&compress}{natbib}
\usepackage[main, final]{neurips_2026}

\usepackage[utf8]{inputenc}
\usepackage[T1]{fontenc}
\usepackage{hyperref}
\usepackage{url}
\usepackage{booktabs}
\usepackage{amsfonts}
\usepackage{nicefrac}
\usepackage{microtype}
\usepackage{xcolor}

\usepackage{graphicx}
\usepackage{placeins}
\usepackage{enumitem}
\usepackage{multirow}
\usepackage{makecell}
\usepackage{wrapfig}
\usepackage{pifont}
\usepackage{array}
\usepackage{colortbl}
\usepackage{xspace}
\usepackage{amsmath}
\usepackage{mathtools}

\definecolor{rankblue}{RGB}{0,174,239}
\definecolor{mitred}{RGB}{163,31,52}

\newcommand{\rankbox}[2]{{\setlength{\fboxsep}{1.5pt}\smash{\colorbox{#1}{#2}}}}
\newcommand{\firstcap}[1]{\rankbox{rankblue!31}{#1}}
\newcommand{\secondcap}[1]{\rankbox{rankblue!16}{#1}}
\newcommand{\thirdcap}[1]{\rankbox{rankblue!8}{#1}}

\newcommand{\cmark}{\ding{51}}
\newcommand{\xmark}{\ding{55}}

\newcommand{\name}{RigidFormer\xspace}

\title{\name: Learning Rigid Dynamics using Transformers}

\author{
  \textbf{Zhiyang Dou}$^{1}$ \quad \textbf{Minghao Guo}$^{1}$ \quad \textbf{Haixu Wu}$^{1}$ \quad \textbf{Doug Roble}$^{2}$ \\[3pt]
  \textbf{Tuur Stuyck}$^{2}$ \quad \textbf{Wojciech Matusik}$^{1}$ \\
  [8pt]
  $^{1}$\,MIT \qquad $^{2}$\,Meta
}

\begin{document}

\maketitle

\input{secs/0_abstract}

\input{secs/1_intro}
\input{secs/2_related}
\input{secs/3_method}
\input{secs/4_exp}
\input{secs/5_conclusion}

\clearpage
\begin{ack}
This work was conducted within the framework of the collaboration between MIT and Meta and was supported by funding from MIT and Meta.
\end{ack}

\bibliographystyle{plainnat}
\bibliography{references}

\appendix
\input{secs/6_appendix}

\end{document}

%% file: secs/0_abstract.tex
\begin{abstract}
Learning-based simulation of multi-object rigid-body dynamics remains difficult because contact is discontinuous and errors compound over long horizons. Most existing methods remain tied to mesh connectivity and vertex-level message passing, which limits their applicability to mesh-free inputs such as point clouds and leads to high computational cost.
Efficiently modeling high-fidelity rigid-body dynamics from mesh-free representations therefore remains challenging. We introduce \textbf{\name}, an object-centric Transformer-based model that learns mesh-free rigid-body dynamics with controllable integration step sizes. \name reasons at the \textit{object level} and advances each object through compact anchors; \textit{Anchor-Vertex Pooling} enriches these anchors with local vertex features, retaining contact-relevant geometry without dense vertex-level interaction. We propose \textit{Anchor-based RoPE} to make attention aware of the objects' spatial layout while respecting the unordered nature of objects and anchors: object-token processing is permutation-equivariant, and the mean-pooled anchor descriptor is invariant to anchor reindexing. \name further enforces \textit{rigidity} by projecting updates onto the rigid-body manifold using differentiable Kabsch alignment. On standard benchmarks, \name outperforms or matches mesh-based baselines using point inputs, runs faster, generalizes to unseen point resolutions and across datasets, and scales to 200+ objects; we also show a preliminary extension to command-conditioned articulated bodies by treating body parts as interacting object-level components. Project page: {\hypersetup{pdfborder={0 0 0}}\href{https://frank-zy-dou.github.io/projects/RigidFormer/}{\textcolor{mitred}{frank-zy-dou.github.io/projects/RigidFormer}}}
\end{abstract}
\vspace{-2.5mm}

%% file: secs/1_intro.tex
\section{Introduction}
\label{sec:intro}

Rigid-body dynamics arises throughout robotics, graphics, and embodied AI. With accurate meshes, reliable physical parameters, and well-tuned contact models, classical physics engines~\cite{coumans2016pybullet,todorov2012mujoco,makoviychuk2021isaac,macklin2022warp,freeman2021brax} can produce faithful trajectories. In practice, however, these prerequisites are often missing: objects may only be available as imperfect or incomplete geometry (e.g., polygon soups or point clouds), with contact properties that are difficult to calibrate. This motivates \emph{mesh-free} modeling. A common choice is a point-based representation, which is easy to acquire, topology-free, and resolution-flexible, making it a natural interface between perception and dynamic scene modeling~\cite{qi2017pointnet,qi2017pointnet++,zhao2021point,lei2025mosca,wang2023tracking,zhang2024monst3r,chen2022virtual}. It also integrates naturally with modern generative pipelines~\cite{vahdat2022lion,yang2019pointflow,huang2026pointworld,zhen2025tesseract}. However, despite the appeal of point-based inputs, many state-of-the-art learned simulators remain mesh-dependent~\cite{pfaff2020learning,allen2022learning,rubanova2024learning,wei2025integrating,yu2023learning}, requiring explicit edge and face connectivity that is not available for point inputs. Moreover, since they typically operate at the vertex, edge, or facet level, their computational costs grow rapidly with resolution, significantly limiting inference efficiency.

We present \name, an object-centric Transformer-based model that learns mesh-free multi-object rigid-body dynamics. Our design is guided by three observations.
First, a rigid body responds to an impulse as a coherent whole: interaction effects do not need to ``diffuse'' across surface vertices edge by edge, as in vertex-centric simulators based on local message passing~\cite{pfaff2020learning,allen2022learning,rubanova2024learning}, which can introduce substantial computational overhead and slow down inference. \name therefore adopts an object-centric representation that reasons over objects rather than vertices: it takes object-level point clouds as input,
\begin{wrapfigure}{r}{68mm}
\vspace{-5mm}
  \hspace*{-3mm}
  \centerline{
  \includegraphics[width=68mm]{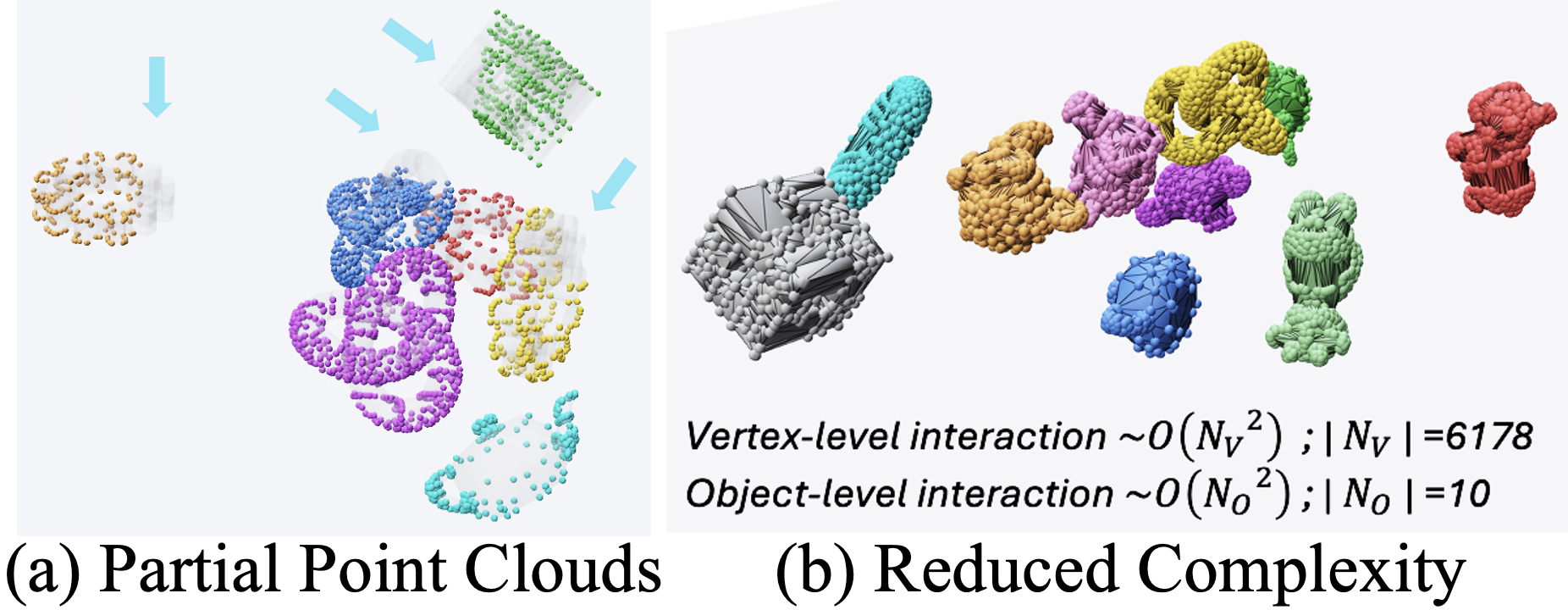}}
  \vspace*{-5mm}
\end{wrapfigure} even partial ones~(see the inset and Sec.~\ref{sec:partial}), and encodes each object into a compact token without requiring connectivity. Interactions are then modeled primarily among \textit{object tokens}, matching the rigid-body assumption that each object moves as a coherent whole. This shift greatly improves efficiency, e.g., 83.6 FPS versus 3.0 FPS, while maintaining simulation quality. We implement this design with Transformers~\cite{vaswani2017attention}, whose attention mechanism flexibly captures multi-object interactions without relying on hand-designed graphs.

Second, we exploit the low-dimensional structure of rigid-body motion during state advancement. Although an object may contain thousands of points, rigid motion lives in a low-dimensional space (6-DoF per object). Therefore, we solve for each object's state update using a small number of \emph{Anchors}, which enables efficient and geometry-aware dynamics updates. Anchor/keypoint and $SE(3)$ representations have precedent in pose tracking and manipulation models~\cite{wang20206pack,byravan2017se3nets}; here, anchors serve as learned simulation states for long-horizon contact dynamics rather than tracked category keypoints or directly regressed part transforms. Given the central role of positional embeddings in Transformer generalization~\cite{su2024roformer,heo2024rotary,zeng2025renderformer}, we propose an \textit{Anchor-based Rotary Positional Embedding~(ARoPE)}~(Sec.~\ref{sec:anchor_rope}) to make attention aware of the objects' spatial layout. Its symmetry properties are deliberately scoped: the object Transformer remains equivariant to object-token permutations because no sequence-index positional embeddings are used, while the mean-pooled ARoPE descriptor is invariant to anchor reindexing within each object. A distance-kernel \textit{Anchor-Vertex Pooling} module further supplies local contact geometry with vertex-order-invariant aggregation. Rather than directly regressing rotations and translations, which can be error-prone~\cite{zhou2019continuity}, we predict anchor motions and obtain the rigid transform via \textit{differentiable rigid projection}, which projects updates onto the rigid-body manifold while preserving gradient flow, improving long-horizon stability. \begin{wraptable}{r}{0.45\linewidth}
\vspace{-6mm}
\centering
\caption{\textbf{Method comparison.} Mesh-free: no triangle connectivity required. Variable step: handles multiple $\Delta t$. Preproc.-free: no offline geometry computation. Warmup frames: number of input frames required for rollout.}
\label{tab:method_comparison}
\vspace{1.5mm}
\setlength{\tabcolsep}{1pt}
\renewcommand{\arraystretch}{1.0}
\scriptsize
\resizebox{\linewidth}{!}{%
\begin{tabular}{%
l
>{\centering\arraybackslash}p{0.14\linewidth}
>{\centering\arraybackslash}p{0.14\linewidth}
>{\centering\arraybackslash}p{0.16\linewidth}
>{\centering\arraybackslash}p{0.16\linewidth}
>{\centering\arraybackslash}p{0.16\linewidth}
}
\toprule
\textbf{Method} & \textbf{Mesh-Free} & \textbf{Var.\ $\Delta t$} & \textbf{Preproc.-Free} & \textbf{\#Warmup Frames $\downarrow$} & \textbf{Runtime (FPS) $\uparrow$}\\
\midrule
MGN     & \xmark & \xmark & \cmark & \textbf{2} & 5.7 \\
FIGNet  & \xmark & \xmark & \cmark & 3 & 3.0 \\
SDF-Sim & \xmark & \xmark & \xmark$^\dagger$  & 3 & -- \\
HopNet  & \xmark & \xmark & \xmark$^\ddagger$ & 3 & 0.2 \\
\midrule
Ours    & \cmark & \cmark & \cmark & \textbf{2} & \textbf{83.6} \\
\bottomrule
\end{tabular}
}
\vspace{-2mm}
\begin{flushleft}
\scriptsize
$^\dagger$ SDF pre-learning: $\sim$5 hours for MOVi-B.\\
$^\ddagger$ Simplicial-complex construction: $\sim$15 days on MOVi-B.
\end{flushleft}
\vspace{-8mm}
\end{wraptable}
Finally, inspired by time-conditioned neural simulation~\cite{yuan2024egode,zhong2021extending,chen2020learning}, \name conditions on the temporal discretization, enabling a single model to operate across step sizes (Sec.~\ref{sec:main_exp}). Larger $\Delta t$ improves long-horizon accuracy by reducing autoregressive error accumulation, while smaller $\Delta t$ captures finer temporal detail when needed.

As a result, \name offers an efficient, stable, and scalable framework for rigid-body dynamics modeling; a comparison with previous methods is in Tab.~\ref{tab:method_comparison}. On MOVi~\cite{greff2022kubric}, it matches or surpasses prior mesh-based methods using only point positions, without requiring mesh connectivity. It leverages known material parameters when available, generalizes to unseen point resolutions and across datasets, supports step-size control, and scales beyond 200 objects while maintaining both accuracy and high inference speed. \name also handles partial point-cloud inputs (see Fig.~\ref{supp_fig:more_partial_vis}). We further show a preliminary application of the same object-anchor design to controllable articulated bodies by treating body parts as interacting object-level components. We summarize our contributions below:
\vspace{-2mm}
\begin{itemize}[leftmargin=*,itemsep=1pt,parsep=0pt,topsep=2pt]
  \item We introduce an efficient and scalable mesh-free Transformer-based neural simulator named \name for multi-object rigid-body dynamics from point representations, supporting simulation across different time-step sizes.
  \item We propose an object-level formulation with Anchor-based RoPE for position-aware attention with explicit object-token permutation equivariance and anchor-order invariance, along with vertex-order-invariant Anchor-Vertex Pooling and a low-dimensional anchor state advance that reduces complexity; rigidity and long-horizon stability are enforced via projection onto the rigid-body manifold during the simulation.
\item We validate \name in diverse experiments, showing fast inference, generalization, scalability, partial point-cloud input, and a preliminary application to command-conditioned articulated bodies.
\end{itemize}
\vspace{-5mm}

%% file: secs/2_related.tex
\section{Related Work}
\label{sec:related}
\begin{figure*}[t]
\centering
\vspace{-7mm}
\includegraphics[width=0.98\linewidth]{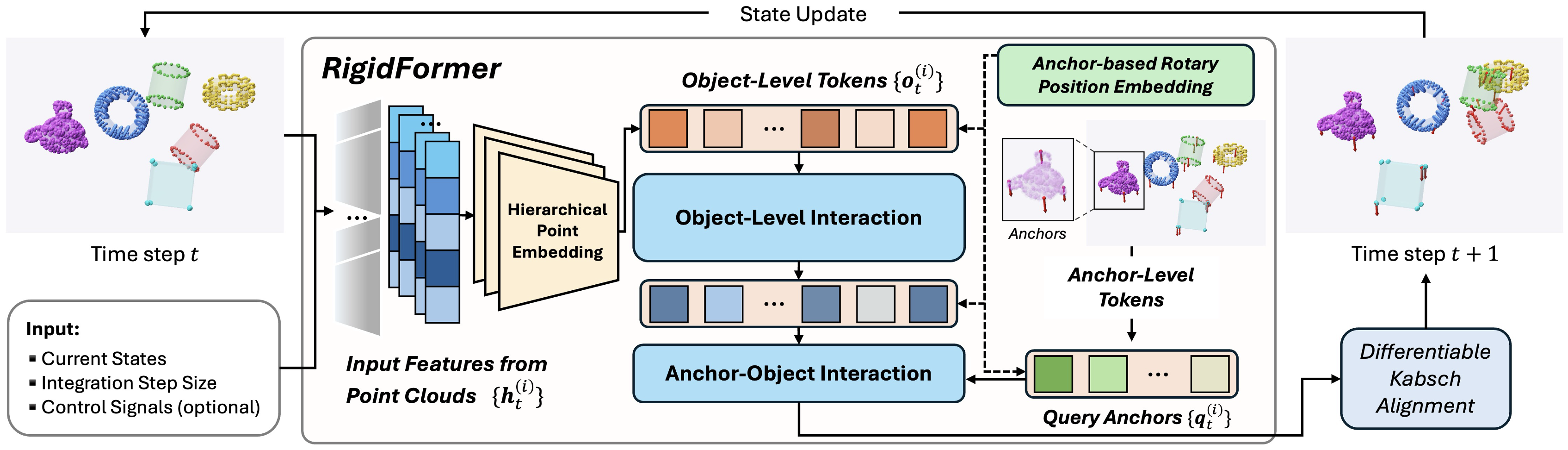}
\vspace{-3mm}
\caption{\textbf{\name Pipeline.} Inputs include two recent point-level states, the time-step size, and optional control signals. At time $t$, we encode each object point cloud into an object token. \emph{Object-Level Interaction} models direct object--object effects, while \emph{Anchor-Object Interaction} lets anchors attend to multi-scale object features and cross-object context. Each object's state is advanced efficiently using a compact set of anchors. We integrate anchor dynamics using Verlet integration followed by differentiable Kabsch alignment to enforce rigidity and obtain the state at $t{+}1$. Anchor-based RoPE is used to improve generalization across object counts.}

\label{fig:pipeline}
\vspace{-4mm}
\end{figure*}

Classical numerical rigid-body simulators~\cite{authors2024genesis, todorov2012mujoco, makoviychuk2021isaac, coumans2016pybullet} resolve contact by solving constrained optimization or complementarity problems.
Differentiable simulators (e.g., DiffTaichi~\cite{hu2019difftaichi}, Warp~\cite{macklin2022warp}, and Brax~\cite{freeman2021brax}) enable gradient-based learning and inverse problems, but they rely on explicit physics engines and typically assume mesh-based geometry rather than mesh-free point inputs.

Early learning-based dynamics models often targeted relatively simple systems with explicit, low-dimensional state representations, typically in 2D. Interaction Networks~\cite{battaglia2016interaction} and Neural Physics Engine~\cite{chang2016compositional} established object- and relation-centric inductive biases and motivated graph-based simulators for dynamics modeling. In rigid-body systems, state-of-the-art neural simulators typically rely on mesh-based inputs in order to faithfully capture the dynamics~\cite{pfaff2020learning,allen2022learning,rubanova2024learning,wei2025integrating}. MeshGraphNets~\cite{pfaff2020learning} extend message passing to mesh discretizations and achieve strong performance for mesh-based simulation. FIGNet~\cite{allen2022learning} improves collision modeling by constructing interactions over mesh faces rather than nodes. HopNet~\cite{wei2025integrating} incorporates higher-order topology and physics-informed message passing for rigid interactions; however, obtaining the required topological structures can be expensive. HCMT~\cite{yu2023learning} uses hierarchical mesh structures and Transformer-style long-range modeling for collision-induced dependencies in flexible-body collision dynamics in the 2D domain. All aforementioned methods require mesh connectivity and incur substantial vertex-level interaction cost as resolution grows. SDF-Sim~\cite{rubanova2024learning} represents shapes with learned signed distance functions, reducing collision-handling bottlenecks but requiring additional shape learning. Compared to these works, \name is mesh-free: it models rigid-body dynamics using point inputs, shifts interaction reasoning to the object level, and uses an anchor-based advance to avoid dense vertex interactions at inference time; see Tab.~\ref{tab:method_comparison}. Rigid-motion and keypoint-based representations have also appeared in related robotics settings. SE3-Nets~\cite{byravan2017se3nets} predict rigid $SE(3)$ transforms for object parts from point clouds and action inputs, demonstrating the value of rigid-motion inductive bias for manipulation. 6-PACK~\cite{wang20206pack} learns anchor-based 3D keypoints for category-level 6D pose tracking. These works are complementary to ours: their keypoints are used for pose estimation in manipulation or tracking, whereas ours are for more dynamic simulation states that are advanced by learned dynamics, coupled with ARoPE and differentiable rigid projection for long-horizon multi-object rollout.

Point-based representations for dynamics have been explored recently. \citet{kim2024object} propose a hierarchical point-cloud representation with continuous point convolutions to improve contact accuracy. \citet{whitney2023learning,whitney2024modeling} learn point-based dynamics by disentangling visual observations from physical states, but accuracy degrades in contact-rich regimes due to the coupling. In contrast, \name adopts an object-level Transformer that effectively models inter-object interactions for points, leading to higher-quality dynamics prediction.

%% file: secs/3_method.tex
\section{Methodology}
\label{sec:method}

We consider a system of $M$ rigid objects, where $M$ can vary across scenes during both training and inference.
Object $i$ is represented by a point set $\mathbf{x}_t^{(i)}\in\mathbb{R}^{N_v^{(i)}\times d}$ at time $t$, where $N_v^{(i)}$ is the number of points and $d$ denotes the per-point feature dimension, and the full state is $\mathbf{x}_t=\{\mathbf{x}_t^{(i)}\}_{i=1}^M$.
We learn a model $f_\theta$ that updates the next state from two consecutive observations:
$
\mathbf{x}_{t+1}=f_\theta(\mathbf{x}_{t-1},\mathbf{x}_t, \Delta t)$. An overview of our pipeline is in Fig.~\ref{fig:pipeline}.

\subsection{Object-Centric Interaction Modeling}

We concatenate, for each vertex of object $i$: (1) the nearest-neighbor displacement vector $\mathbf{d}_t^{(i)}\!\in\!\mathbb{R}^{N_v^{(i)}\times 3}$ from the vertex to the nearest point on another object or the ground plane, (2) the velocity $\mathbf{v}_t^{(i)}=\mathbf{x}_t^{(i)}-\mathbf{x}_{t-1}^{(i)}\!\in\!\mathbb{R}^{N_v^{(i)}\times 3}$, (3) the reference-offset $\mathbf{r}_t^{(i)}=\mathbf{x}_t^{(i)}-\mathbf{x}_{\mathrm{ref}}^{(i)}\!\in\!\mathbb{R}^{N_v^{(i)}\times 3}$ where the reference is the first frame of the trajectory, and (4) physics parameters $\boldsymbol{\phi}^{(i)}=[m,\mu,\epsilon]\!\in\!\mathbb{R}^{3}$ (mass, friction, restitution). These yield the input feature $\mathbf{h}_t^{(i)}\!\in\!\mathbb{R}^{N_v^{(i)}\times 12}$. Inspired by PointNet~\cite{qi2017pointnet}, we build an encoder $\mathrm{Enc}_\theta$ with hierarchical feature extraction to aggregate per-vertex features into a fixed-dimensional object embedding: $\mathbf{o}_t^{(i)} = \mathrm{Enc}_\theta(\mathbf{h}_t^{(i)}) \in \mathbb{R}^{D}$. After computing per-vertex features, we extract multi-scale geometry at the global level and three subsampled levels; these features are then concatenated and fused into one object-level embedding. This design captures both fine-grained local geometry and coarse global structure while remaining robust to variable vertex counts (see validation in Sec.~\ref{sec:main_exp}). The encoder is \emph{shared} across all objects, promoting generalization to various geometries.
Using one token per object drastically shortens the Transformer sequence compared with vertex-level modeling, substantially improving efficiency; see Appendix~\ref{supp_sec:runtime}.

Given object embeddings $\mathbf{O}_t=[\mathbf{o}_t^{(1)},\ldots,\mathbf{o}_t^{(M)}]\in\mathbb{R}^{M\times D}$, our decoder stacks $L$ Transformer blocks initialized from $\mathbf{O}_t$ and $N_r{=}16$ learned register tokens. For clarity, we describe the object-token update and omit the registers from notation. At layer $\ell$, the decoder applies residual self-attention, step-size FiLM conditioning~\cite{perez2018film}, and a residual feed-forward update:
$\tilde{\mathbf{Z}}_t^{(\ell)}=\mathrm{SelfAttn}(\mathbf{Z}_t^{(\ell-1)})+\mathbf{Z}_t^{(\ell-1)}$,
$\hat{\mathbf{Z}}_t^{(\ell)}=\boldsymbol{\gamma}_\ell(\mathbf{c})\odot\tilde{\mathbf{Z}}_t^{(\ell)}+\boldsymbol{\beta}_\ell(\mathbf{c})$,
and
$\mathbf{Z}_t^{(\ell)}=\mathrm{FFN}(\hat{\mathbf{Z}}_t^{(\ell)})+\hat{\mathbf{Z}}_t^{(\ell)}$,
for $\ell=1,\ldots,L$. The FiLM code $\mathbf{c}=(s,s^2)\in\mathbb{R}^2$ encodes the integration step size, where $s\propto\Delta t$ captures first-order time scaling and $s^2$ mirrors the $\Delta t^2$ factor in Verlet integration. The layer-specific MLPs $\boldsymbol{\gamma}_\ell$ and $\boldsymbol{\beta}_\ell$ produce channel-wise scale and shift parameters, allowing the same decoder to adapt its features across different temporal discretizations. Motivated by gated attention in language modeling~\cite{qiu2025gated}, we modulate each attention update with a query-conditioned sigmoid gate:
$\mathbf{y}=\sigma(\mathbf{G}(\mathbf{Q}))\odot \mathrm{Attn}(\mathbf{Q},\mathbf{K},\mathbf{V})$,
where $\mathbf{G}(\mathbf{Q})$ has the same per-head channel shape as the attention output. In our setting, the gate acts as a learned attenuator for noisy or weakly relevant interaction reads, which stabilizes autoregressive dynamics rollouts and improves long-horizon accuracy (Sec.~\ref{sec:ablation_study}). The output of this stage is a set of updated object-level tokens $\{\mathbf{Z}_t^{(L)}\}$ that summarize scene context and object interactions. We next use anchors as queries into these object tokens, combine the retrieved context with local vertex features, and advance the system through anchor dynamics.

\subsection{Anchor-based State Advance}
Rigid-body motion is low-dimensional (6-DoF) even when an object contains thousands of points. Therefore, directly updating all vertices is expensive and redundant: dense attention over $MN_v$ points costs $O((MN_v)^2)$, and per-vertex regression destabilizes prediction. Meanwhile, regressing rotations and translations directly~\cite{zhou2019continuity} can be error-prone and unstable due to discontinuities in common $SE(3)$ parameterizations. In \name, we instead select a small set of $N_a{=}4$ \textit{anchors} per object using FPS~\cite{gonzalez1985clustering,qi2017pointnet++}, reducing the interaction cost to $O((MN_a)^2)$ with $N_a \ll N_v$ across time steps. We form anchor queries by extracting features at anchor locations and projecting them with an MLP to obtain $\mathbf{Q}_t \in \mathbb{R}^{(MN_a)\times D}$. Each anchor query attends to the updated object tokens $\{\mathbf{Z}_t^{(L)}\}$ via cross-attention, retrieving object-level interaction context, and the network then predicts a per-anchor acceleration $\mathbf{a}_t^{(i,k)}$.

Anchor queries summarize a rigid object's dynamics, but accurate acceleration prediction during contact depends strongly on which vertices lie close to the contact site. To inject this fine-grained, contact-local geometry into each anchor without paying the cost of full per-vertex attention, we attach an Anchor-Vertex Pooling~(AVP) module that aggregates per-vertex encoder features around each anchor with a learnable isotropic distance kernel:
\begingroup
\setlength{\abovedisplayskip}{4pt}
\setlength{\belowdisplayskip}{4pt}
\begin{equation}
\mathbf{u}_t^{(i,k)} = \frac{\sum_{v=1}^{N_v^{(i)}} w_t^{(i,k,v)}\,\mathbf{f}_t^{(i,v)}}{\sum_{v=1}^{N_v^{(i)}} w_t^{(i,k,v)}}, \quad
w_t^{(i,k,v)} = \exp\!\left(-\frac{\|\mathbf{x}_t^{(i,v)}-\mathbf{q}_t^{(i,k)}\|}{\sigma}\right),
\end{equation}
\endgroup
where $\mathbf{f}_t^{(i,v)}$ is the encoder feature at vertex $v$, $\mathbf{q}_t^{(i,k)}$ is the position of anchor $k$, and the kernel bandwidth $\sigma$ is trained jointly with the rest of the network; padded vertices are masked in the batched implementation. Because the weights depend only on point-anchor distances and the aggregation is a normalized sum, AVP is invariant to vertex ordering, and its attention weights are unchanged by a common rigid transform of the point and anchor coordinates. The resulting feature $\mathbf{u}_t^{(i,k)}\!\in\!\mathbb{R}^{256}$ is concatenated to the anchor query before predicting acceleration, enriching it with collision-aware local context. We advance anchors with Verlet integration to obtain candidate anchor positions from predicted accelerations:
$
\hat{\mathbf{q}}_{t+1}^{(i,k)}=\mathbf{a}_t^{(i,k)}\Delta t^2+2\mathbf{q}_t^{(i,k)}-\mathbf{q}_{t-1}^{(i,k)}$.

\noindent\textbf{Scatter to All Vertices via Differentiable Rigid Projection.} We recover the rigid transform $(\mathbf{R}^{(i)},\mathbf{t}^{(i)})\in SE(3)$ by aligning reference anchors $\mathbf{q}_{\mathrm{ref}}^{(i,k)}$ to the candidate anchors $\hat{\mathbf{q}}_{t+1}^{(i,k)}$ using Kabsch alignment~\cite{kabsch1976solution}:
\begingroup
\setlength{\abovedisplayskip}{4pt}
\setlength{\belowdisplayskip}{1pt}
\begin{equation}\begin{aligned}
\mathbf{H} &= \sum_k (\mathbf{q}_{\text{ref}}^{(i,k)} - \bar{\mathbf{q}}_{\text{ref}}^{(i)})(\hat{\mathbf{q}}_{t+1}^{(i,k)} - \bar{\hat{\mathbf{q}}}^{(i)})^\top,
\quad \mathbf{U}, \mathbf{\Sigma}, \mathbf{V}^\top = \operatorname{SVD}(\mathbf{H}), \\
\mathbf{R}^{(i)} &= \mathbf{V} \operatorname{diag}(1, 1, \det(\mathbf{V}\mathbf{U}^\top)) \mathbf{U}^\top,
\quad \mathbf{t}^{(i)} = \bar{\hat{\mathbf{q}}}^{(i)} - \mathbf{R}^{(i)} \bar{\mathbf{q}}_{\text{ref}}^{(i)},
\end{aligned}\end{equation}
\endgroup
where $k$ indexes anchors, and $\bar{\hat{\mathbf{q}}}^{(i)}$ and $\bar{\mathbf{q}}_{\text{ref}}^{(i)}$ denote centroids of the predicted and reference anchor sets. Then, we update the full-resolution point set by broadcasting the transform to all vertices: $\mathbf{x}_{t+1}^{(i,v)}=\mathbf{R}^{(i)}\mathbf{x}_{\mathrm{ref}}^{(i,v)}+\mathbf{t}^{(i)}, \quad \forall v\in\{1,\ldots,N_v\}$. This projection enforces rigidity by construction and improves long-horizon rollout stability. Since gradients through SVD can be unstable near degenerate singular values, we implement rigid registration with RoMa~\cite{bregier2021deep} for robust differentiability.

\subsection{Anchor-based Rotary Positional Embedding}
\label{sec:anchor_rope}
Effective positional embeddings are essential for the Transformer-based \name to generalize rigid-body dynamics, where interactions vary with object count and 3D geometry. In \name, object tokens have no inherent ordering---permuting objects should not change the predicted dynamics. Moreover, contact outcomes depend on relative 3D geometry rather than absolute indices. Encoding all vertices in attention is largely redundant, hurts generalization, and adds prohibitive computational overhead.

We propose \textit{Anchor-based Rotary Positional Embedding (ARoPE)}, which encodes the spatial layout of objects from a sparse set of anchor positions, making the attention position-aware while remaining efficient and generalizable across different numbers of objects with various shapes. Concretely, for object $i$ with anchors $\{\mathbf{x}^{(i)}_k\}_{k=1}^{N_a}\!\subset\!\mathbb{R}^3$ ($N_a{=}4$), we apply a shared 3D rotary anchor map $\psi_\omega(\cdot)$ to each anchor, mapping each coordinate through $d_c{=}32$ rotary phase channels to obtain a per-anchor 96-D phase descriptor. We then aggregate the per-anchor descriptors by mean-pooling:
\begingroup
\setlength{\abovedisplayskip}{4pt}
\setlength{\belowdisplayskip}{4pt}
\begin{equation}
\displaystyle
\mathrm{ARoPE}(\{\mathbf{x}^{(i)}_k\}) =
\frac{1}{N_a}\sum_{k=1}^{N_a}
\underbrace{\bigoplus_{j=1}^{3}\bigoplus_{l=1}^{d_c/2}\!
\left[\omega_l x^{(i)}_{k,j},\,\omega_l x^{(i)}_{k,j}\right]}_{\psi_\omega(\mathbf{x}^{(i)}_k)} ,
\end{equation}
\endgroup
where $\omega_l$ are log-spaced frequencies, $\oplus$ denotes concatenation, and the repeated terms form the even--odd channel pairs used by RoPE. The resulting descriptor provides the RoPE angles used in attention. For an attention head with query/key channels split into a rotary part and a pass-through part, $\mathbf{q}=[\mathbf{q}_r;\mathbf{q}_p]$ and $\mathbf{k}=[\mathbf{k}_r;\mathbf{k}_p]$, ARoPE applies
\begin{equation}
\tilde{\mathbf{q}}=[\mathbf{q}_r\odot\cos\mathbf{a}_q+\operatorname{rot}(\mathbf{q}_r)\odot\sin\mathbf{a}_q;\mathbf{q}_p],\quad
\tilde{\mathbf{k}}=[\mathbf{k}_r\odot\cos\mathbf{a}_k+\operatorname{rot}(\mathbf{k}_r)\odot\sin\mathbf{a}_k;\mathbf{k}_p],
\end{equation}
where $\mathbf{a}_q$ and $\mathbf{a}_k$ are the ARoPE descriptors for the query and key tokens, and $\operatorname{rot}(\cdot)$ swaps each even--odd channel pair with a sign flip as in standard RoPE~\cite{su2024roformer}. Mean-pooling these per-anchor rotary features---rather than concatenating raw anchor coordinates as in a naive multi-point variant---matches the symmetry that anchor identities are arbitrary. Since the rotary angles are linear in the coordinates, the mean-pooled descriptor equals the RoPE angles of the anchor centroid: ARoPE encodes where each object is, while its shape reaches attention through the object tokens and the AVP features. ARoPE is invariant to anchor reindexing: for any anchor permutation $\pi$, $\mathrm{ARoPE}(\{\mathbf{x}^{(i)}_{\pi(k)}\}_{k=1}^{N_a})
= \frac{1}{N_a}\sum_{k=1}^{N_a}\psi_\omega(\mathbf{x}^{(i)}_{\pi(k)})
= \mathrm{ARoPE}(\{\mathbf{x}^{(i)}_k\}_{k=1}^{N_a})$, because the sum is unchanged by reordering. Applying ARoPE to object tokens yields improved performance and generalization across varying object counts (Sec.~\ref{sec:ablation_study}). Detailed proofs and discussion are given in Appendix~\ref{supp_sec:theoretical_properties}.

\subsection{Training Objectives} \label{sec:training_obj}

Our objective combines position and acceleration L1 losses: $
\mathcal{L} = \lambda_{\text{pos}}(\mathcal{L}_{\text{pos}}^{\text{raw}} + \mathcal{L}_{\text{pos}}^{\text{rigid}}) + \lambda_{\text{acc}}(\mathcal{L}_{\text{acc}}^{\text{raw}} + \mathcal{L}_{\text{acc}}^{\text{rigid}})
$, where ``raw'' and ``rigid'' denote losses computed before and after Kabsch alignment, respectively, with $\lambda_{\text{pos}}{=}10$ and $\lambda_{\text{acc}}{=}1$, applied to a Smooth L1 reduction~\cite{girshick2015fast}. All four terms are supervised at the selected anchors; full-resolution vertices are deterministic after the rigid projection. We train \name on NVIDIA A100 GPUs for $300$ epochs with AdamW~\cite{loshchilov2019decoupled} ($\beta_1{=}0.9$, $\beta_2{=}0.999$, weight decay $0.01$) at a base learning rate of $10^{-4}$. The schedule combines a $10$-epoch linear warmup (from $10\%$ of the base rate) with cosine decay to $\eta_{\min}{=}10^{-6}$, and we clip the gradient norm at $1.0$ for stability. Additional implementation details are provided in Appendices~\ref{supp_section:training_details} and \ref{supp_sec:architecture}.

%% file: secs/4_exp.tex
\begin{figure*}[t]
\vspace{-2mm}
    \centering
    \includegraphics[width=\linewidth]{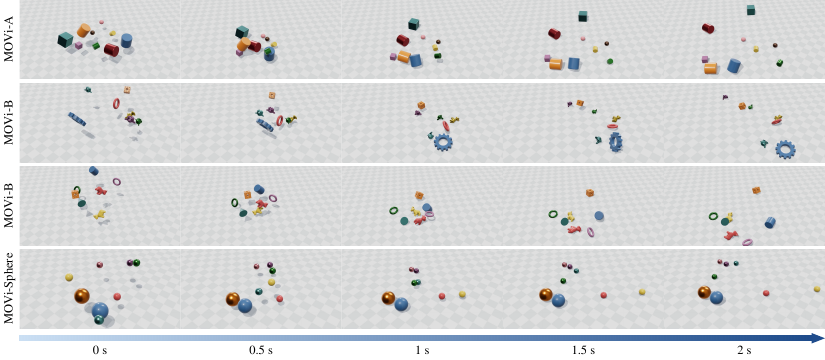}
    \vspace{-5mm}
\caption{\textbf{Qualitative results from \name} on test scenes of MOVi-A, MOVi-B and MOVi-Sphere, from the initial state (0\,s) to 2\,s at step size 1. Meshes are shown only for visualization; our model operates on point inputs. Additional visualization results on the MOVi datasets are provided in Appendix~\ref{supp_sec:more_vis}.}
    \label{fig:main_visualization}
    \vspace{-5mm}
\end{figure*}

\section{Experiments}
\label{sec:exp}

We conduct experiments to evaluate \name: (i) \textit{Accuracy} vs. state-of-the-art learning-based simulators; (ii) \textit{Generalization} across datasets; (iii) \textit{Resolution generalization} to unseen test-time point counts; (iv) \textit{Anchor robustness} to varying anchor counts and anchor perturbations (random sampling); (v) \textit{Ablations} of anchor-based 3D RoPE, gated attention, and differentiable rigid projection; (vi) \textit{Scalability and controllability} for large scenes and articulated bodies; (vii) \textit{Runtime performance}; and (viii) \textit{Partial point clouds}.
Please refer to the \textit{project page} for more qualitative results.

\noindent\textbf{Datasets \& Metrics.} We use the MOVi (Multi-Object Video) datasets~\cite{greff2022kubric}:
\textit{MOVi-A} (basic geometric shapes), \textit{MOVi-B} (complex geometric shapes), and \textit{MOVi-Sphere} (spheres). Following~\cite{allen2022learning,wei2025integrating,rubanova2024learning}, we report Translation RMSE (m) for per-object center-of-mass position and Orientation RMSE (deg) via quaternion geodesic distance.
We evaluate at physical frames during autoregressive rollout, following the HopNet protocol~\cite{wei2025integrating}.
For our variable step sizes (1/5/10), predictions are mapped to the corresponding physical frames for fair comparison.
All metrics are averaged over the test set. We use MOVi-B for ablations due to its diversity and vertex-level variation (see statistics in Appendix~\ref{supp_section:data_statistics}).

\subsection{Main Comparison}
\label{sec:main_exp}

\begin{table*}[t]
\vspace{-7mm}
\centering
\caption{\textbf{Standard Performance on MOVi-A, MOVi-B, and MOVi-Sphere.} Each cell reports position RMSE (m)\,/\,orientation RMSE ($^{\circ}$) at prediction horizons of 50, 75, and 100 frames. Per-column top-three ranks are denoted by shading: \firstcap{1st}, \secondcap{2nd}, \thirdcap{3rd}.}
\label{tab:main_comparison}
\vspace{1mm}
\setlength{\tabcolsep}{1.5pt}
\renewcommand{\arraystretch}{1.1}
\newcommand{\mF}[1]{\rankbox{rankblue!31}{#1}}
\newcommand{\mS}[1]{\rankbox{rankblue!16}{#1}}
\newcommand{\mT}[1]{\rankbox{rankblue!8}{#1}}
\resizebox{\textwidth}{!}{%
\begin{tabular}{lccccccccccc}
\toprule
\multirow{2}{*}{\textbf{Model}} & \multicolumn{3}{c}{\textbf{MOVi-A}} & & \multicolumn{3}{c}{\textbf{MOVi-B}} & & \multicolumn{3}{c}{\textbf{MOVi-Sphere}} \\
\cmidrule(lr){2-4}\cmidrule(lr){6-8}\cmidrule(lr){10-12}
& \textbf{50} & \textbf{75} & \textbf{100} & & \textbf{50} & \textbf{75} & \textbf{100} & & \textbf{50} & \textbf{75} & \textbf{100} \\
\midrule
MGN+ & 0.705\,/\,31.21 & N/A & N/A &  & 0.538\,/\,26.91 & N/A & N/A &  & N/A & N/A & N/A \\
MGN-LargeRadius+ & 0.119\,/\,15.07 & N/A & N/A &  & 0.460\,/\,26.34 & N/A & N/A &  & N/A & N/A & N/A \\
FIGNet & \mT{0.115}\,/\,14.84 & N/A & N/A &  & \mT{0.127}\,/\,13.99 & N/A & N/A &  & N/A & N/A & N/A \\
FIGNet$_{reimpl}$ & 0.132\,/\,7.10 & \mT{0.285}\,/\,14.62 & \mT{0.492}\,/\,23.30 &  & 0.141\,/\,7.39 & \mT{0.300}\,/\,15.16 & \mT{0.516}\,/\,24.96 &  & N/A & N/A & N/A \\
HCMT$_{reimpl}$$^{*}$ & 0.239\,/\,5.70 & 0.538\,/\,\mT{11.82} & 0.951\,/\,\mS{18.40} &  & 0.237\,/\,\mT{4.72} & 0.527\,/\,\mT{9.80} & 0.932\,/\,\mT{17.43} &  & \mT{0.243}\,/\,\mT{4.19} & 0.541\,/\,\mS{8.21} & 0.956\,/\,\mT{13.81} \\
VPD$^{\dagger}_{reimpl}$ & 0.235\,/\,\mS{5.10} & 0.489\,/\,\mS{11.66} & 0.827\,/\,20.37 &  & 0.275\,/\,\mS{4.65} & 0.581\,/\,\mS{9.70} & 0.987\,/\,\mS{16.99} &  & 0.244\,/\,4.47 & \mT{0.510}\,/\,\mT{9.50} & \mT{0.855}\,/\,17.65 \\
HopNet & \mS{0.054}\,/\,\mT{5.64} & \mS{0.115}\,/\,11.84 & \mS{0.196}\,/\,\mT{18.83} &  & \mS{0.047}\,/\,4.91 & \mS{0.101}\,/\,10.35 & \mS{0.176}\,/\,17.91 &  & \mS{0.034}\,/\,\mS{4.05} & \mS{0.073}\,/\,\mS{8.21} & \mS{0.124}\,/\,\mS{13.68} \\
\midrule
\name$^{\dagger,*}$ & \mF{0.040}\,/\,\mF{4.81} & \mF{0.086}\,/\,\mF{10.64} & \mF{0.145}\,/\,\mF{17.57} &  & \mF{0.030}\,/\,\mF{3.29} & \mF{0.061}\,/\,\mF{7.16} & \mF{0.108}\,/\,\mF{12.65} &  & \mF{0.022}\,/\,\mF{2.93} & \mF{0.045}\,/\,\mF{6.28} & \mF{0.078}\,/\,\mF{10.74} \\
\bottomrule
\multicolumn{12}{l}{\footnotesize $*$ Transformer-based model; $^{\dagger}$ Point inputs.} \\
\end{tabular}
}
\vspace{-6mm}
\end{table*}
\begin{table*}[t]
  \centering
  \caption{\textbf{Generalization performance across MOVi-Sphere (S), MOVi-A (A), and MOVi-B (B).} The left block compares FIGNet, HopNet, and \name under the matched step-size-1 setting, with top-three rank shading: \firstcap{1st}, \secondcap{2nd}, \thirdcap{3rd}. The right block reports \name with larger step sizes; step size 10 predicts only frames that are multiples of 10 (--).}
  \label{tab:generalization}
  \vspace{1mm}
  \newcommand{\mF}[1]{\rankbox{rankblue!31}{#1}}
  \newcommand{\mS}[1]{\rankbox{rankblue!16}{#1}}
  \newcommand{\mT}[1]{\rankbox{rankblue!8}{#1}}
\begin{minipage}[t]{0.62\textwidth}
\centering
\setlength{\tabcolsep}{1.2pt}
\small
\resizebox{\linewidth}{!}{%
\begin{tabular}{llcc@{\hspace{5pt}}cc@{\hspace{5pt}}cc}
  \toprule
  \multirow{2}{*}{\textbf{Train}} & \multirow{2}{*}{\textbf{Test}} &
  \multicolumn{2}{c}{\textbf{FIGNet}} &
  \multicolumn{2}{c}{\textbf{HopNet}} &
  \multicolumn{2}{c}{\textbf{\name{} step=1}} \\
  \cmidrule(lr){3-4} \cmidrule(lr){5-6} \cmidrule(lr){7-8}
   &  & \textbf{75} & \textbf{100} & \textbf{75} & \textbf{100} & \textbf{75} & \textbf{100} \\
  \midrule
  \multirow{2}{*}{S}
   & A & \mT{0.370}\,/\,\mT{14.08} & \mT{0.626}\,/\,\mT{22.46} & \mF{0.112}\,/\,\mS{11.29} & \mF{0.197}\,/\,\mS{17.74} & \mS{0.122}\,/\,\mF{10.79} & \mS{0.205}\,/\,\mF{17.57} \\
   & B & \mT{0.662}\,/\,\mT{18.08} & \mT{1.149}\,/\,\mT{29.66} & \mS{0.106}\,/\,\mS{9.75} & \mS{0.188}\,/\,\mS{17.13} & \mF{0.081}\,/\,\mF{8.56} & \mF{0.138}\,/\,\mF{15.02} \\
  \midrule
  \multirow{2}{*}{A}
   & S & \mT{0.239}\,/\,\mT{11.55} & \mT{0.417}\,/\,\mT{19.66} & \mS{0.100}\,/\,\mS{9.16} & \mS{0.172}\,/\,\mS{15.17} & \mF{0.059}\,/\,\mF{7.17} & \mF{0.100}\,/\,\mF{12.59} \\
   & B & \mT{0.512}\,/\,\mT{13.49} & \mT{0.864}\,/\,\mT{22.41} & \mS{0.117}\,/\,\mS{10.77} & \mS{0.202}\,/\,\mS{18.66} & \mF{0.074}\,/\,\mF{8.07} & \mF{0.132}\,/\,\mF{14.46} \\
  \midrule
  \multirow{2}{*}{B}
   & S & \mT{0.536}\,/\,\mT{12.74} & \mT{0.905}\,/\,\mT{20.42} & \mS{0.095}\,/\,\mS{9.09} & \mS{0.160}\,/\,\mS{15.04} & \mF{0.063}\,/\,\mF{7.78} & \mF{0.110}\,/\,\mF{13.43} \\
   & A & \mT{0.520}\,/\,\mT{14.53} & \mT{0.871}\,/\,\mT{22.89} & \mS{0.120}\,/\,\mS{12.08} & \mS{0.202}\,/\,\mS{19.15} & \mF{0.099}\,/\,\mF{10.62} & \mF{0.165}\,/\,\mF{17.18} \\
  \bottomrule
\end{tabular}}
\end{minipage}
\hfill
\begin{minipage}[t]{0.37\textwidth}
\centering
\setlength{\tabcolsep}{1.2pt}
\small
\resizebox{\linewidth}{!}{%
\begin{tabular}{llcccc}
  \toprule
  \multirow{2}{*}{\textbf{Train}} & \multirow{2}{*}{\textbf{Test}} &
  \multicolumn{2}{c}{\textbf{\name{} step=5}} &
  \multicolumn{2}{c}{\textbf{\name{} step=10}} \\
  \cmidrule(lr){3-4} \cmidrule(lr){5-6}
   &  & \textbf{75} & \textbf{100} & \textbf{75} & \textbf{100} \\
  \midrule
  \multirow{2}{*}{S}
   & A & 0.099\,/\,8.32 & 0.178\,/\,14.80 & -- & 0.139\,/\,11.37 \\
   & B & 0.067\,/\,7.10 & 0.123\,/\,13.31 & -- & 0.094\,/\,10.04 \\
  \midrule
  \multirow{2}{*}{A}
   & S & 0.046\,/\,5.46 & 0.083\,/\,10.34 & -- & 0.074\,/\,7.58 \\
   & B & 0.057\,/\,6.66 & 0.108\,/\,12.40 & -- & 0.085\,/\,9.77 \\
  \midrule
  \multirow{2}{*}{B}
   & S & 0.053\,/\,6.24 & 0.095\,/\,11.64 & -- & 0.082\,/\,8.67 \\
   & A & 0.078\,/\,8.20 & 0.134\,/\,14.37 & -- & 0.108\,/\,11.63 \\
  \bottomrule
\end{tabular}}
\end{minipage}
  \vspace{-6mm}
\end{table*}

Tab.~\ref{tab:main_comparison} reports the matched step-size-1 setting on MOVi benchmarks. We compare \name (point input, $^\dagger$) against mesh-based simulators (e.g., HopNet, FIGNet, MGN), point-based methods (VPD), and Transformer-based approaches (HCMT). Despite using no mesh connectivity, \name obtains the best position and orientation errors in every reported column. On MOVi-B at 100 frames, \name improves over HopNet from $0.176$/ $17.91^\circ$ to $0.108$/ $12.65^\circ$, while using only point inputs. For fairness, the input points are the mesh vertices themselves, as used by mesh-based methods, even when sparse for simple shapes (e.g., 51 points for a cube, 64 for a cone); Sec.~\ref{sec:partial} uses partial point clouds. We run VPD~\cite{whitney2023learning} with at most 1{,}536 points per scene (Appendix~\ref{supp_sec:training}). The SDF-Sim~\cite{rubanova2024learning} code is not publicly available, so we cannot rerun it under our full protocol; in the MOVi-B @50 result reported by Rubanova et al., SDF-Sim obtains $0.160$/ $18.03^\circ$ (their FIGNet: $0.140$/ $15.00^\circ$), whereas \name obtains $0.030$/ $3.29^\circ$ at matched step size $1$ and $0.012$/ $1.22^\circ$ at step size $10$, without SDF fitting. Qualitative results are provided in Fig.~\ref{fig:main_visualization} and on the project page. We evaluate temporal discretization through step-size-conditioned rollouts in Tab.~\ref{tab:ablation_steps}.

\noindent
\textbf{Cross-Dataset Generalization.} Tab.~\ref{tab:generalization} reports cross-dataset transfer by training on one MOVi variant and testing on another. In the matched step-size-1 block, \name improves over FIGNet in every split and over HopNet in every split except the position error from MOVi-Sphere to MOVi-A. The larger-step block shows that step-size conditioning further reduces our long-horizon errors in every transfer split. Training on MOVi-Sphere transfers least well in position; we analyze this trend in Appendix~\ref{supp_sec:cross_dataset_analysis}.

\begin{wraptable}{r}{0.5\columnwidth}
\vspace{-4mm}
\centering
\caption{\textbf{MOVi-C, zero-shot}: position (m)\,/\,orientation ($^{\circ}$) RMSE at 50 frames. HopNet as reported in~\cite{wei2025integrating}. Step size 10: given frames spaced by the step.}
\label{tab:movic}
\vspace{1mm}
\newcommand{\mF}[1]{\rankbox{rankblue!31}{#1}}
\resizebox{\linewidth}{!}{%
\begin{tabular}{lccc}
\toprule
& \multicolumn{2}{c}{\textbf{Step size 1}} & \textbf{Step size 10} \\
\cmidrule(lr){2-3}\cmidrule(lr){4-4}
\textbf{Trained on} & HopNet & \name & \name \\
\midrule
MOVi-Sphere & 0.104\,/\,\mF{6.56} & \mF{0.096}\,/\,7.55 & 0.032\,/\,2.26 \\
MOVi-A & 0.109\,/\,6.96 & \mF{0.095}\,/\,\mF{6.95} & 0.032\,/\,2.12 \\
MOVi-B & 0.105\,/\,7.09 & \mF{0.086}\,/\,\mF{5.88} & 0.032\,/\,2.02 \\
\bottomrule
\end{tabular}}
\vspace{-0.2mm}
\end{wraptable}
\noindent
\textbf{Zero-Shot Transfer to MOVi-C.} Following HopNet~\cite{wei2025integrating}, we evaluate the models trained on MOVi-Sphere, MOVi-A and MOVi-B without adaptation on the 120 MOVi-C test scenes of HopNet, which hold 3 to 10 scanned everyday objects~\cite{downs2022google} unlike the training data in shape, size and mass (Appendix~\ref{supp_sec:movic}). At step size 1 and 50 frames, \name has 8--18\% lower position errors than HopNet from all training sets (Tab.~\ref{tab:movic}); its orientation error is lower from MOVi-B, nearly equal from MOVi-A and higher from MOVi-Sphere. Step size 10 lowers its errors further.

\noindent
\textbf{Point Resolution Generalization.}
Tab.~\ref{tab:moviB_pointcloud_res} reports point-cloud results on MOVi-B. \name is trained with randomly sampled point counts $\{128,256,512,1024\}$ and evaluated at 768 points. Despite the unseen test-time resolution, the model remains stable across 50/75/100-frame rollouts\footnote{In Tabs.~\ref{tab:moviB_pointcloud_res}--\ref{tab:random_anchors}, the 75-frame entries of step size $10$ are evaluated at frame 80, the nearest frame on its grid.}; at 100 frames, step sizes $10$, $5$, and $1$ obtain $0.137$/$11.13^\circ$, $0.161$/$14.83^\circ$, and $0.189$/$16.22^\circ$, respectively.

\begin{table*}[t]
\centering
\vspace{-7mm}
\caption{\textbf{Step-size-conditioned rollout performance.} Each cell reports position RMSE (m)\,/\,orientation RMSE ($^{\circ}$). Step size 10 predicts only frames that are multiples of 10 (--).}
\label{tab:ablation_steps}
\vspace{1mm}
\setlength{\tabcolsep}{6pt}
\renewcommand{\arraystretch}{1.1}
\resizebox{\textwidth}{!}{%
\begin{tabular}{cccccccccccc}
\toprule
\multirow{2}{*}{\textbf{Step Size}} & \multicolumn{3}{c}{\textbf{MOVi-A}} & & \multicolumn{3}{c}{\textbf{MOVi-B}} & & \multicolumn{3}{c}{\textbf{MOVi-Sphere}} \\
\cmidrule(lr){2-4}\cmidrule(lr){6-8}\cmidrule(lr){10-12}
& \textbf{50} & \textbf{75} & \textbf{100} & & \textbf{50} & \textbf{75} & \textbf{100} & & \textbf{50} & \textbf{75} & \textbf{100} \\
\midrule
10 & 0.016\,/\,1.43 & -- & 0.092\,/\,11.26 && 0.012\,/\,1.22 & -- & 0.071\,/\,8.74 && 0.010\,/\,0.88 & -- & 0.062\,/\,6.58 \\
5  & 0.029\,/\,2.94 & 0.068\,/\,7.76 & 0.120\,/\,13.76 && 0.020\,/\,2.71 & 0.048\,/\,6.32 & 0.090\,/\,11.52 && 0.017\,/\,2.02 & 0.040\,/\,4.83 & 0.073\,/\,8.85 \\
1  & 0.040\,/\,4.81 & 0.086\,/\,10.64 & 0.145\,/\,17.57 && 0.030\,/\,3.29 & 0.061\,/\,7.16 & 0.108\,/\,12.65 && 0.022\,/\,2.93 & 0.045\,/\,6.28 & 0.078\,/\,10.74 \\
\bottomrule
\end{tabular}
}
\vspace{-6mm}
\end{table*}

\begin{table*}[t]
\centering

\begin{minipage}[t]{0.35\textwidth}
\centering
\caption{\textbf{Generalization to different point resolutions.} Trained with point counts 128, 256, 512, 1024 and evaluated at 768 points. Each cell reports position RMSE (m)\,/\,orientation RMSE ($^{\circ}$).}
\label{tab:moviB_pointcloud_res}
\vspace{1mm}
\setlength{\tabcolsep}{4pt}
\small
\resizebox{\linewidth}{!}{%
\begin{tabular}{c ccc}
\toprule
\multirow{2}{*}{\textbf{Step size}} &
\multicolumn{3}{c}{\textbf{Rollout horizon (frames)}} \\
\cmidrule(lr){2-4}
& \textbf{50} & \textbf{75} & \textbf{100} \\
\midrule
10 & 0.071\,/\,1.64 & 0.101\,/\,6.11 & 0.137\,/\,11.13 \\
5  & 0.082\,/\,3.55 & 0.113\,/\,8.06 & 0.161\,/\,14.83 \\
1  & 0.094\,/\,4.38 & 0.132\,/\,9.21 & 0.189\,/\,16.22 \\
\bottomrule
\end{tabular}
}
\end{minipage}
\hfill
\begin{minipage}[t]{0.62\textwidth}
\centering
\vspace{2mm}

\caption{\textbf{Effect of Gated Attention and Differentiable Rigid Alignment.} Top-three values shaded by rank among variants.}
\label{tab:ablation_combined_gated_diff}
\vspace{1mm}
\setlength{\tabcolsep}{1pt}
\renewcommand{\arraystretch}{1.05}
\small
\newcommand{\gF}[1]{\rankbox{rankblue!31}{#1}}
\newcommand{\gS}[1]{\rankbox{rankblue!16}{#1}}
\newcommand{\gT}[1]{\rankbox{rankblue!8}{#1}}
\resizebox{\linewidth}{!}{%
\begin{tabular}{cccccccccccc}
\toprule
\multirow{2}{*}{\makecell[c]{\textbf{Gated}\\ \textbf{Attention}}} &
\multirow{2}{*}{\makecell[c]{\textbf{Differentiable}\\ \textbf{Alignment}}} &
\multirow{2}{*}{\textbf{Metric}} &
\multicolumn{3}{c}{\textbf{step\_size=1}} & \multicolumn{3}{c}{\textbf{step\_size=5}} & \multicolumn{3}{c}{\textbf{step\_size=10}} \\
\cmidrule(lr){4-6} \cmidrule(lr){7-9} \cmidrule(lr){10-12}
 &  &  & \textbf{50} & \textbf{75} & \textbf{100} & \textbf{50} & \textbf{75} & \textbf{100} & \textbf{50} & \textbf{75} & \textbf{100} \\
\midrule
\multirow{2}{*}{\ding{55}}
 & \multirow{2}{*}{\ding{51}}
 & Pos & \gT{0.077} & \gT{0.153} & \gT{0.259} & \gT{0.050} & \gT{0.107} & \gT{0.191} & \gT{0.033} & \gT{0.090} & \gT{0.152} \\
 &  & Ori & \gT{4.70} & \gT{9.98} & \gT{17.12} & \gT{3.51} & \gT{7.77} & \gT{13.93} & \gT{1.57} & \gT{6.02} & \gF{10.78} \\
\midrule
\multirow{2}{*}{\ding{51}}
 & \multirow{2}{*}{\ding{55}}
 & Pos & \gS{0.052} & \gS{0.098} & \gS{0.169} & \gS{0.042} & \gS{0.082} & \gS{0.146} & \gS{0.030} & \gS{0.072} & \gS{0.121} \\
 &  & Ori & \gS{4.08} & \gS{8.79} & \gS{15.79} & \gS{3.14} & \gS{7.39} & \gS{13.75} & \gS{1.55} & \gS{5.98} & \gT{11.04} \\
\midrule
\multirow{2}{*}{\ding{51}}
 & \multirow{2}{*}{\ding{51}}
 & Pos & \gF{0.050} & \gF{0.095} & \gF{0.161} & \gF{0.040} & \gF{0.078} & \gF{0.136} & \gF{0.029} & \gF{0.069} & \gF{0.115} \\
 &  & Ori & \gF{3.97} & \gF{8.51} & \gF{15.33} & \gF{3.06} & \gF{7.25} & \gF{13.55} & \gF{1.51} & \gF{5.89} & \gS{10.85} \\
\bottomrule
\end{tabular}
}
\end{minipage}
\vspace{-6mm}
\end{table*}

\noindent
\textbf{Step Sizes.}
We investigate the effect of step-size conditioning in Tab.~\ref{tab:ablation_steps}. Larger step sizes consistently yield better long-horizon performance because the model performs fewer autoregressive updates over the same physical horizon. Step size $=10$ gives the lowest 100-frame errors on MOVi-A, MOVi-B, and MOVi-Sphere. The step size $=1$ rows remain the matched setting for comparison with prior one-step rollout protocols, while step size $=5$ provides an intermediate trade-off between rollout frequency and local prediction difficulty. More discussion can be found in Appendix~\ref{supp_sec:step_size_discussion}.

\subsection{Ablation Studies}
\label{sec:ablation_study}

\noindent
\textbf{Positional Embedding.}
We compare our Anchor-based Rotary Positional Embedding (ARoPE) with standard sinusoidal PE~\cite{vaswani2017attention}, learned absolute PE~\cite{devlin2019bert}, and geometry-aware OBB, PCA, and SE(3) baselines. ARoPE uses the set-aggregated anchor descriptor from Sec.~\ref{sec:anchor_rope} to rotate query and key vectors as a function of sparse 3D anchor coordinates, while mean aggregation makes the descriptor invariant to anchor reindexing. As reported in Tab.~\ref{tab:ablation_pe}, ARoPE achieves the best or tied-best position error in eight of nine cells and the best orientation error in most cells; SE(3) and sinusoidal PE are close in selected metrics. Overall, the results suggest that direct anchor-coordinate RoPE provides a stable positional signal for heterogeneous MOVi-B shapes.

\noindent
\textbf{Gated Attention.} We ablate the sigmoid gating in \name. Tab.~\ref{tab:ablation_combined_gated_diff} shows that gating mainly improves translation accuracy and long-horizon stability. At 100 steps, it reduces position error from $0.259\!\rightarrow\!0.161$ (step\_size$=1$), $0.191\!\rightarrow\!0.136$ (step\_size$=5$), and $0.152\!\rightarrow\!0.115$ (step\_size$=10$). Orientation error also improves in most cells; at step size $10$ and 100 steps, it remains essentially tied with the ungated variant.

\noindent
\textbf{Differentiable Kabsch Alignment.} Tab.~\ref{tab:ablation_combined_gated_diff} also compares \name \textbf{w/} vs.\ \textbf{w/o} differentiable rigid projection. With gating enabled, the differentiable variant consistently reduces 100-step position error ($0.169\!\rightarrow\!0.161$, $0.146\!\rightarrow\!0.136$, and $0.121\!\rightarrow\!0.115$ for step sizes 1, 5, and 10). Orientation error also decreases at the same horizons, indicating that gradient flow through the rigid projection reduces rollout drift.

\begin{wraptable}{r}{0.62\columnwidth}
\vspace{-6mm}
\centering
\caption{\textbf{Comparison of Different Positional Embeddings.} Top-three values shaded by rank among PE variants.}
\label{tab:ablation_pe}
\vspace{1mm}
\setlength{\tabcolsep}{2pt}
\renewcommand{\arraystretch}{0.9}
\small
\newcommand{\pF}[1]{\rankbox{rankblue!31}{#1}}
\newcommand{\pS}[1]{\rankbox{rankblue!16}{#1}}
\newcommand{\pT}[1]{\rankbox{rankblue!8}{#1}}
\resizebox{\linewidth}{!}{%
\begin{tabular}{cccccccccccc}
\toprule
\multirow{2}{*}{\makecell[c]{\textbf{PE}}} & \multirow{2}{*}{\textbf{Metric}} &
\multicolumn{3}{c}{\textbf{step\_size=1}} & \multicolumn{3}{c}{\textbf{step\_size=5}} & \multicolumn{3}{c}{\textbf{step\_size=10}} \\
\cmidrule(lr){3-5} \cmidrule(lr){6-8} \cmidrule(lr){9-11}
 &  & \textbf{50} & \textbf{75} & \textbf{100} & \textbf{50} & \textbf{75} & \textbf{100} & \textbf{50} & \textbf{75} & \textbf{100} \\
\midrule
\multirow{2}{*}{Sinusoidal}
 & Pos & \pS{0.051} & \pT{0.100} & \pT{0.172} & \pT{0.042} & 0.083 & \pT{0.145} & \pS{0.030} & \pT{0.074} & \pT{0.122} \\
 & Ori & \pS{4.10} & \pS{8.68} & \pF{15.30} & \pT{3.16} & \pS{7.39} & \pF{13.52} & \pT{1.55} & 6.04 & \pS{10.91} \\
\midrule
\multirow{2}{*}{Learned}
 & Pos & 0.054 & 0.103 & 0.176 & \pT{0.042} & \pT{0.082} & 0.146 & \pS{0.030} & \pS{0.073} & \pT{0.122} \\
 & Ori & \pT{4.24} & \pT{9.01} & \pT{15.94} & 3.18 & \pT{7.43} & \pT{13.77} & 1.56 & 5.98 & \pT{11.04} \\
\midrule
\multirow{2}{*}{OBB}
 & Pos & 0.060 & 0.114 & 0.189 & 0.045 & 0.087 & 0.155 & \pT{0.033} & 0.079 & 0.133 \\
 & Ori & 4.38 & 9.44 & 16.83 & 3.31 & 7.69 & 14.35 & 1.57 & \pS{5.96} & 11.08 \\
\midrule
\multirow{2}{*}{PCA}
 & Pos & 0.380 & 0.867 & 1.548 & 0.240 & 0.647 & 1.249 & 0.107 & 0.499 & 0.918 \\
 & Ori & 7.02 & 15.11 & 25.79 & 5.23 & 12.52 & 22.62 & 2.27 & 9.32 & 16.74 \\
\midrule
\multirow{2}{*}{SE(3)}
 & Pos & \pT{0.052} & \pS{0.099} & \pS{0.167} & \pS{0.041} & \pS{0.080} & \pS{0.139} & \pF{0.029} & \pF{0.069} & \pF{0.114} \\
 & Ori & 4.25 & 9.21 & 16.36 & \pS{3.15} & 7.45 & 13.89 & \pS{1.52} & \pT{5.97} & \pT{11.04} \\
\midrule
\multirow{2}{*}{\makecell[c]{ARoPE\\(Ours)}}
 & Pos & \pF{0.050} & \pF{0.095} & \pF{0.161} & \pF{0.040} & \pF{0.078} & \pF{0.136} & \pF{0.029} & \pF{0.069} & \pS{0.115} \\
 & Ori & \pF{3.97} & \pF{8.51} & \pS{15.33} & \pF{3.06} & \pF{7.25} & \pS{13.55} & \pF{1.51} & \pF{5.89} & \pF{10.85} \\
\bottomrule
\end{tabular}
}
\vspace{-4mm}
\end{wraptable}

\begin{figure*}[t]
\vspace{-7mm}
    \centering
    \includegraphics[width=\linewidth]{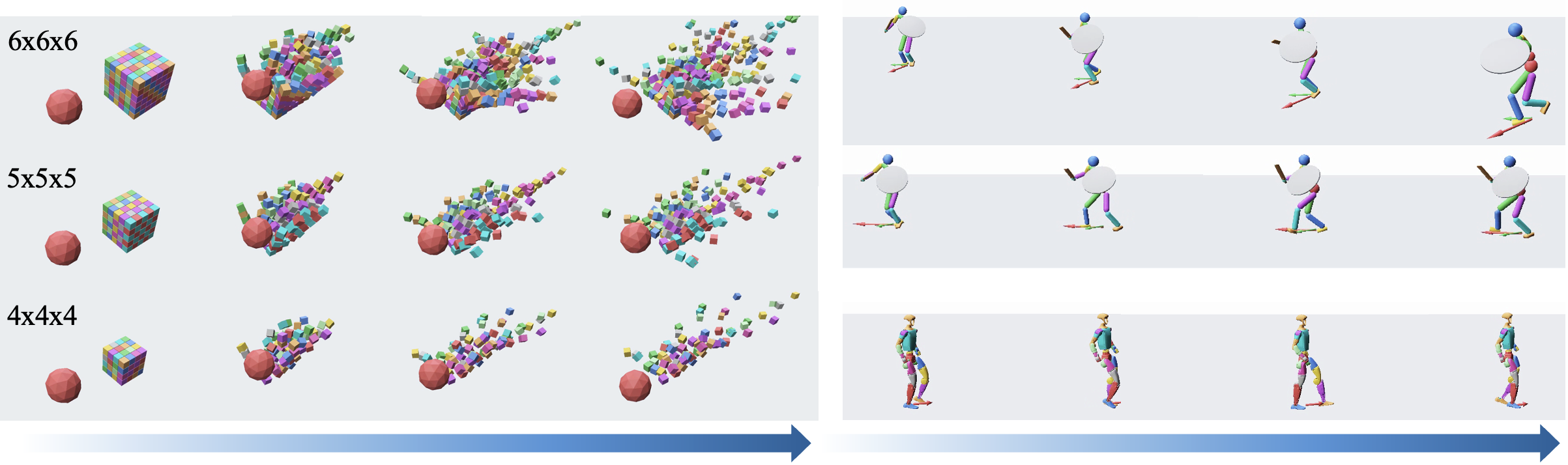}
    \vspace{-7mm}
\caption{\textbf{Scalability and controllability.} \textbf{Left:} Simulation with $216, 125, 64$ cubes; \name stays stable as object count grows. \textbf{Right:} Preliminary direction-controlled articulated dynamics. Humanoid: commanded facing and moving directions are the green and red arrows, respectively; Unitree G1: the arrow shows the moving direction. \name follows heading commands while preserving coherent part-level motion.}
    \label{fig:applications}
    \vspace{-5.5mm}
\end{figure*}

\noindent
\textbf{Different Numbers of Anchors.} Tab.~\ref{tab:diff_numbers_anchors} studies the effect of the anchor count $k$ on MOVi-B. Increasing from $k=3$ to $k=4$ improves translation accuracy across step sizes and rollout horizons. Using $k=8$ remains competitive and yields lower orientation error in several long-horizon settings, but it uses twice as many anchor queries and offers a less favorable translation--cost trade-off. We therefore use 4 anchors as the default efficiency--quality trade-off.

\noindent
\textbf{Randomized FPS Anchors.}
We train and evaluate \name to assess whether the model learns \textit{intrinsic geometric features} of the object, rather than relying on specific anchor identities. Tab.~\ref{tab:random_anchors} reports a perturbation study where FPS anchors are randomly re-sampled during both training and evaluation; it is separate from the default fixed-$N_a{=}4$ evaluation used elsewhere. Randomized FPS with $4$ anchors gives the strongest translation accuracy, while $8$ anchors can reduce rotation error in several settings. These results indicate that \name is robust to anchor selection and that the default $N_a{=}4$ provides a good quality--efficiency trade-off.

\begin{table*}[t]
\centering
\begin{minipage}[t]{0.49\textwidth}
\centering
\caption{\textbf{Different Numbers of Anchors.} Results for different numbers of anchors in \name. Each cell reports position RMSE (m)\,/\,orientation RMSE ($^{\circ}$). Top three shaded, also in Tab.~\ref{tab:random_anchors}.}
\label{tab:diff_numbers_anchors}
\vspace{1.0mm}
\setlength{\tabcolsep}{2.5pt}
\small
\newcommand{\rF}[1]{\rankbox{rankblue!31}{#1}}
\newcommand{\rS}[1]{\rankbox{rankblue!16}{#1}}
\newcommand{\rT}[1]{\rankbox{rankblue!8}{#1}}
\resizebox{\linewidth}{!}{%
\begin{tabular}{cc ccc}
\toprule
\multirow{2}{*}{\textbf{Anchors}} &
\multirow{2}{*}{\textbf{Step size}} &
\multicolumn{3}{c}{\textbf{Rollout horizon (frames)}} \\
\cmidrule(lr){3-5}
& & \textbf{50} & \textbf{75} & \textbf{100} \\
\midrule
\multirow{3}{*}{3}
  & 10 & \rT{0.033}\,/\,\rS{1.63} & \rT{0.081}\,/\,\rT{6.74} & \rT{0.135}\,/\,\rT{12.71} \\
  & 5  & \rT{0.047}\,/\,\rT{3.63} & \rS{0.094}\,/\,\rT{8.67} & \rS{0.166}\,/\,\rT{16.80} \\
  & 1  & \rS{0.060}\,/\,\rT{4.82} & \rS{0.118}\,/\,\rT{10.64} & \rS{0.198}\,/\,\rT{19.01} \\
\midrule
\multirow{3}{*}{\makecell[c]{4\\}}
  & 10 & \rF{0.029}\,/\,\rF{1.51} & \rF{0.069}\,/\,\rS{5.89} & \rF{0.115}\,/\,\rS{10.85} \\
  & 5  & \rF{0.040}\,/\,\rF{3.06} & \rF{0.078}\,/\,\rS{7.25} & \rF{0.136}\,/\,\rS{13.55} \\
  & 1  & \rF{0.050}\,/\,\rF{3.97} & \rF{0.095}\,/\,\rF{8.51} & \rF{0.161}\,/\,\rS{15.33} \\
\midrule
\multirow{3}{*}{8}
  & 10 & \rS{0.031}\,/\,\rF{1.51} & \rS{0.079}\,/\,\rF{5.53} & \rS{0.130}\,/\,\rF{10.07} \\
  & 5  & \rS{0.046}\,/\,\rS{3.14} & \rT{0.097}\,/\,\rF{7.20} & \rT{0.170}\,/\,\rF{13.38} \\
  & 1  & \rT{0.061}\,/\,\rS{4.08} & \rT{0.123}\,/\,\rS{8.67} & \rT{0.208}\,/\,\rF{15.26} \\
\bottomrule
\end{tabular}
}
\end{minipage}
\hfill
\begin{minipage}[t]{0.49\textwidth}
\centering
\caption{\textbf{Randomized FPS anchors.} We train and evaluate with FPS anchors randomly re-sampled as a perturbation study. Each cell reports position RMSE (m)\,/\,orientation RMSE ($^{\circ}$).}
\label{tab:random_anchors}
\vspace{1.0mm}
\setlength{\tabcolsep}{2.5pt}
\small
\newcommand{\rF}[1]{\rankbox{rankblue!31}{#1}}
\newcommand{\rS}[1]{\rankbox{rankblue!16}{#1}}
\newcommand{\rT}[1]{\rankbox{rankblue!8}{#1}}
\resizebox{\linewidth}{!}{%
\begin{tabular}{cc ccc}
\toprule
\multirow{2}{*}{\textbf{Anchors}} &
\multirow{2}{*}{\textbf{Step size}} &
\multicolumn{3}{c}{\textbf{Rollout horizon (frames)}} \\
\cmidrule(lr){3-5}
& & \textbf{50} & \textbf{75} & \textbf{100} \\
\midrule
\multirow{3}{*}{3}
  & 10 & \rT{0.031}\,/\,\rT{1.72} & \rS{0.075}\,/\,\rT{6.36} & \rS{0.126}\,/\,\rT{11.74} \\
  & 5  & \rS{0.042}\,/\,\rT{3.72} & \rS{0.085}\,/\,\rT{8.40} & \rS{0.149}\,/\,\rT{15.12} \\
  & 1  & \rS{0.055}\,/\,\rT{4.80} & \rS{0.110}\,/\,\rT{10.46} & \rS{0.187}\,/\,\rT{18.19} \\
\midrule
\multirow{3}{*}{4}
  & 10 & \rF{0.029}\,/\,\rS{1.59} & \rF{0.070}\,/\,\rS{5.76} & \rF{0.115}\,/\,\rS{10.39} \\
  & 5  & \rF{0.041}\,/\,\rF{3.21} & \rF{0.081}\,/\,\rF{7.23} & \rF{0.139}\,/\,\rS{13.38} \\
  & 1  & \rF{0.052}\,/\,\rS{4.41} & \rF{0.099}\,/\,\rS{9.31} & \rF{0.163}\,/\,\rS{16.16} \\
\midrule
\multirow{3}{*}{8}
  & 10 & \rS{0.030}\,/\,\rF{1.52} & \rT{0.083}\,/\,\rF{5.58} & \rT{0.141}\,/\,\rF{10.06} \\
  & 5  & \rT{0.044}\,/\,\rS{3.23} & \rT{0.095}\,/\,\rS{7.28} & \rT{0.171}\,/\,\rF{13.29} \\
  & 1  & \rT{0.056}\,/\,\rF{4.04} & \rT{0.114}\,/\,\rF{8.64} & \rT{0.193}\,/\,\rF{15.17} \\
\bottomrule
\end{tabular}
}
\end{minipage}
\vspace{-3.17mm}
\end{table*}

\subsection{Scalability, Controllability and Efficiency}
\noindent\textbf{Large-Scale Simulation.}
We demonstrate the scalability of \name on the WreckingBall dataset, which contains scenes with $64$, $125$, and $216$ cubes (visualized in Fig.~\ref{fig:applications}, left). At 50 physical steps, \name achieves position RMSEs of $1.210$, $0.690$, and $0.130$, with orientation errors of $20.50^\circ$, $16.50^\circ$, and $4.60^\circ$ for the three scenes, respectively. The configurations differ in more than object count, so they are stress tests of one model rather than an object-count ablation (App.~\ref{supp_sec:large_scale}); the model remains stable in all of them. In this setting, \name runs at 20 FPS.

\noindent\textbf{Controllable Articulated-Body Simulation.}
We further evaluate \name as a preliminary extension to articulated characters under directional control. For the ASE humanoid with 17 body parts and the Unitree G1 robot with 31 body parts, we treat each body part as an object-level component and condition the model on desired heading commands through FiLM. As shown in Fig.~\ref{fig:applications}, \name produces coherent whole-body motion that follows different heading commands, suggesting that the object-centric formulation can extend beyond independent rigid objects. At 100 physical steps, \name obtains position/orientation errors of $0.062$ / $14.47^\circ$ on ASE Humanoid and $0.072$ / $16.26^\circ$ on G1.
We note that this result is presented as an extension study rather than the primary evidence for the method.

\label{sec:runtime_comp}
\begin{wraptable}{r}{0.32\columnwidth}
\vspace{-7mm}
\centering
\caption{Runtime comparison.}
\vspace{1mm}
\label{tab:runtime_comparison}
\resizebox{\linewidth}{!}{
\begin{tabular}{lcc}
\toprule
Method & ms/step & FPS \\
\midrule
HopNet & 4228.7 & 0.2 \\
FIGNet & 336.0 & 3.0 \\
\textbf{\name (Ours)} & \textbf{12.0} & \textbf{83.6} \\
\bottomrule
\end{tabular}%
}
\vspace{-4mm}
\end{wraptable}
\noindent\textbf{Runtime Performance.} Tab.~\ref{tab:runtime_comparison} reports runtime on MOVi-B using a 50-step autoregressive rollout averaged over 10 iterations on an NVIDIA GeForce RTX 5080. \name achieves $28\times$ and $354\times$ speedups over FIGNet and HopNet, respectively. This efficiency mainly comes from the object-centric Transformer design, which reasons over compact object-level states rather than dense vertex-level structures. Further runtime analysis is provided in Appendix~\ref{supp_sec:runtime}.

\subsection{Partial Point Clouds}
\label{sec:partial}
\begin{figure*}[t]
    \centering
    \includegraphics[width=\linewidth]{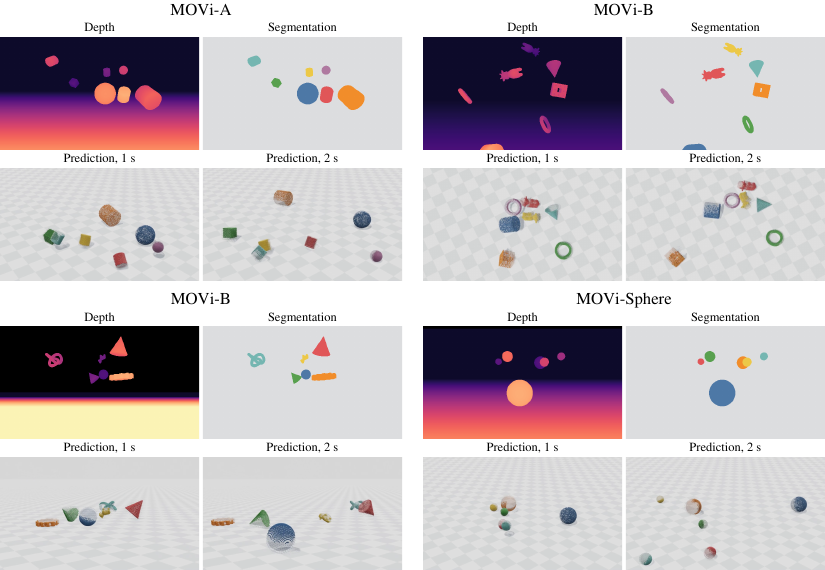}
    \vspace{-5mm}
\caption{\textbf{Partial point clouds.} Test scenes of MOVi-A, MOVi-B and MOVi-Sphere observed by a single depth camera. For each scene: the depth and segmentation images (the inputs) and the predicted rollout after 1\,s and 2\,s. Opaque points are observed; light points are completed.}
    \label{fig:partial_examples}
    \vspace{-3mm}
\end{figure*}
\name also simulates scenes observed by a single depth camera (Fig.~\ref{fig:partial_examples}). The camera observes each scene once, so each object is given as a partial point cloud: the back-projected points of its segment, which cover only the side facing the camera. The model takes these points either directly or after a point completion network fills in the occluded side; the architecture is unchanged, the points take the place of the mesh vertices, and the model is trained on these inputs. Appendix~\ref{supp_sec:partial_depth} describes the camera, the completion network and the training details, and compares both inputs (Tab.~\ref{supp_tab:partial}).

%% file: secs/5_conclusion.tex
\section{Conclusion}
\label{sec:conclusion}
We presented \name, an object-centric, mesh-free Transformer-based simulator for multi-object rigid-body contact dynamics from point clouds. \name shifts interaction reasoning to the \textit{object level} and updates dynamics using a compact set of anchors, reducing vertex-level computation while maintaining accuracy. To encode the objects' spatial layout, we introduce Anchor-based RoPE, whose mean-pooled anchor descriptor is invariant to anchor reindexing and whose use for object tokens preserves permutation equivariance over unordered objects. \name further enforces rigidity by projecting updates onto the rigid-body manifold via differentiable Kabsch alignment, and it supports controllable integration through step-size conditioning within a single model. Experiments validate the efficiency, effectiveness, and versatility of \name, including zero-shot transfer to scanned objects and simulation from single-camera partial point clouds. \noindent \textit{Limitations and future work.}
The current formulation primarily targets object-level rigid-body contact dynamics and relies on object labels to identify which points belong to each object. Although we report partial point-cloud results, prediction becomes challenging when partial observations capture too little of an object's shape. Future work could move toward noisier perception settings, mixed rigid--deformable scenes, and more fine-grained adaptive time stepping (Appendix~\ref{supp_sec:limitations_future}).

%% file: secs/6_appendix.tex
\newpage

\renewcommand\thefigure{\Alph{section}\arabic{figure}}
\renewcommand\thetable{\Alph{section}\arabic{table}}

\section*{Appendix Contents}
\begingroup
\setlength{\parindent}{0pt}
\newcommand{\appcontentsline}[2]{\hyperref[#1]{#2}\dotfill\pageref{#1}\par}
\appcontentsline{supp_sec:more_vis}{A. More Qualitative Results}
\appcontentsline{supp_sec:theoretical_properties}{B. Theoretical Properties of the Object-Anchor Representation}
\appcontentsline{supp_section:training_details}{C. Training Details}
\appcontentsline{supp_section:data_statistics}{D. Data Statistics}
\appcontentsline{supp_sec:augmentation}{E. Data Augmentation}
\appcontentsline{supp_sec:runtime}{F. Runtime Performance and Computational Costs}
\appcontentsline{supp_sec:architecture}{G. Detailed Network Architecture}
\appcontentsline{supp_sec:movic}{H. Zero-Shot Transfer to MOVi-C}
\appcontentsline{supp_sec:partial_depth}{I. Partial Point Clouds from a Single Depth Camera}
\appcontentsline{supp_sec:large_scale}{J. Large-Scale Simulation}
\appcontentsline{supp_sec:articulated}{K. Controllable Articulated Body Simulation}
\appcontentsline{supp_sec:additional}{L. Additional Ablations and Analyses}
\appcontentsline{supp_sec:step_size_discussion}{M. Discussion on Temporal Step Sizes}
\appcontentsline{supp_sec:limitations_future}{N. Limitations, Future Work, and Impact}
\endgroup
\vspace{1mm}
\clearpage

\section{More Qualitative Results}
\label{supp_sec:more_vis}

\begin{figure}[h]
    \centering
    \includegraphics[width=\linewidth]{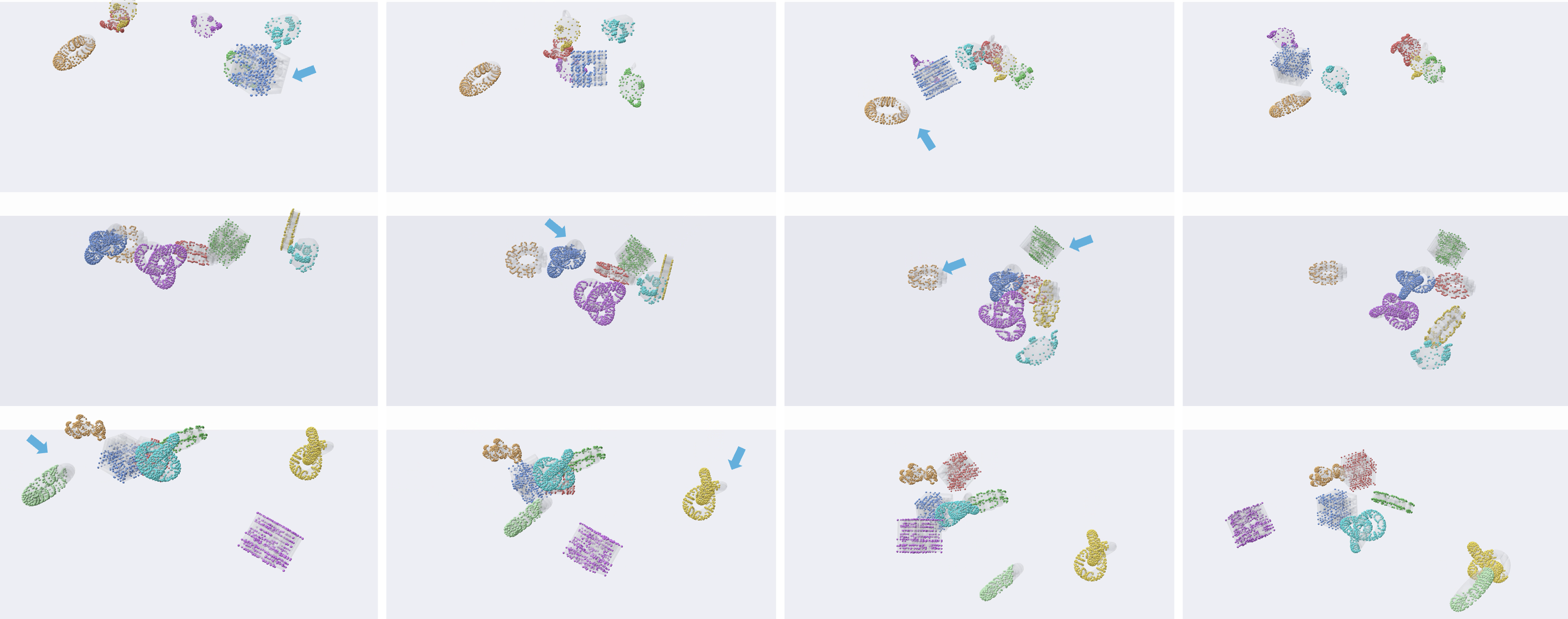}
    \vspace{-4mm}
\caption{\textbf{Partial point-cloud qualitative results.} \name predicts accurate rigid-body interactions from partial observations with low drift and stable contacts. Meshes are shown only for visualization; our model operates on point inputs. Please refer to the project page for more results.}
    \label{supp_fig:more_partial_vis}
\end{figure}

More results on partial point-cloud inputs are shown in Fig.~\ref{supp_fig:more_partial_vis}. We randomly mask 25\% of the points inside each object's bounding box and evaluate the model from the resulting partial observations. Additional videos are included on the project page.

\begin{figure}[h]
\vspace{-2mm}
    \centering
    \includegraphics[width=\linewidth]{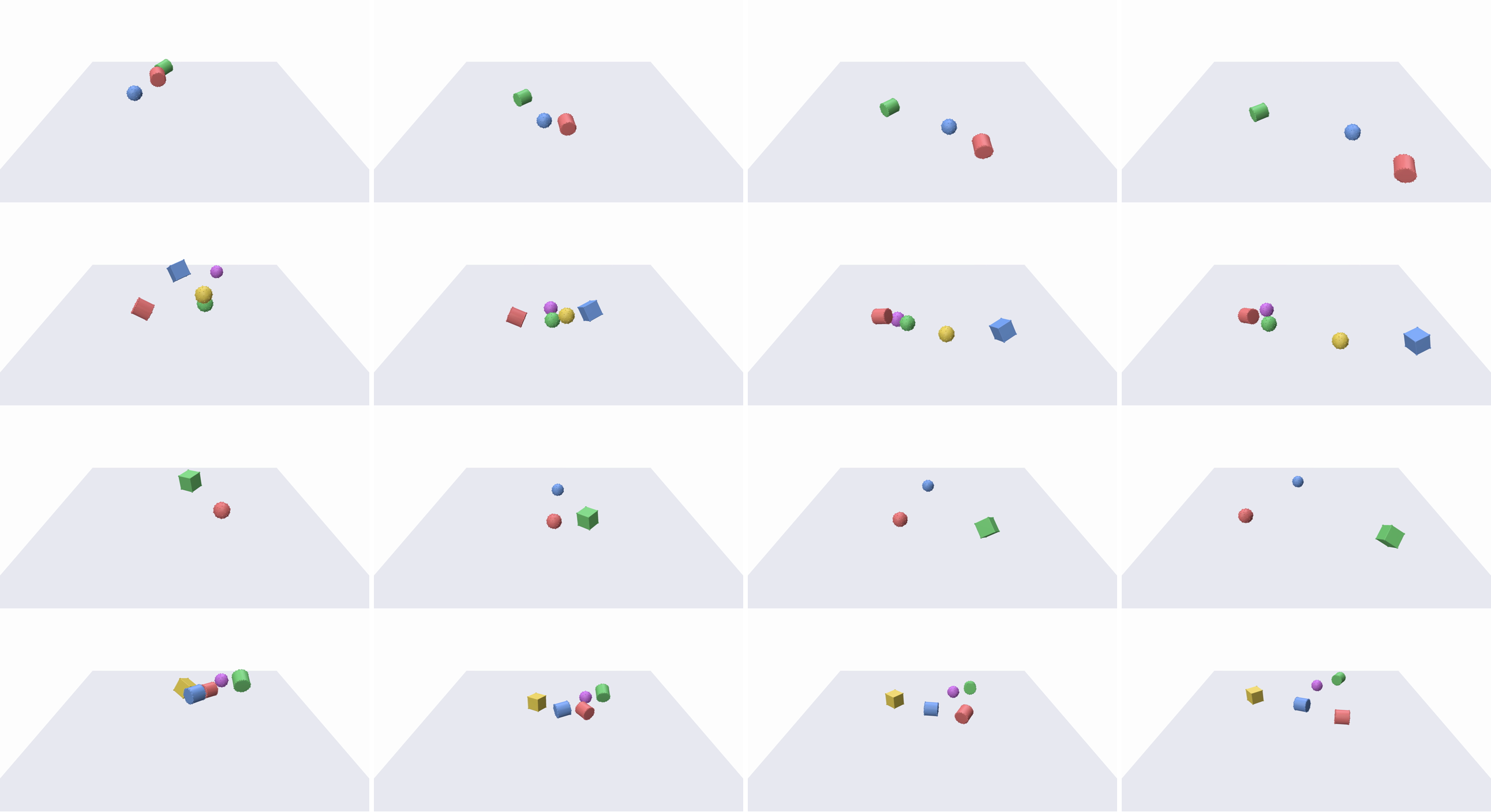}
    \vspace{-4mm}
\caption{\textbf{More qualitative results on MOVi-A.} Each row is one held-out test sample; columns show four frames from the rollout.}
    \label{supp_fig:more_vis_a}
\end{figure}

\begin{figure}
\vspace{-7mm}
    \centering
    \includegraphics[width=\linewidth]{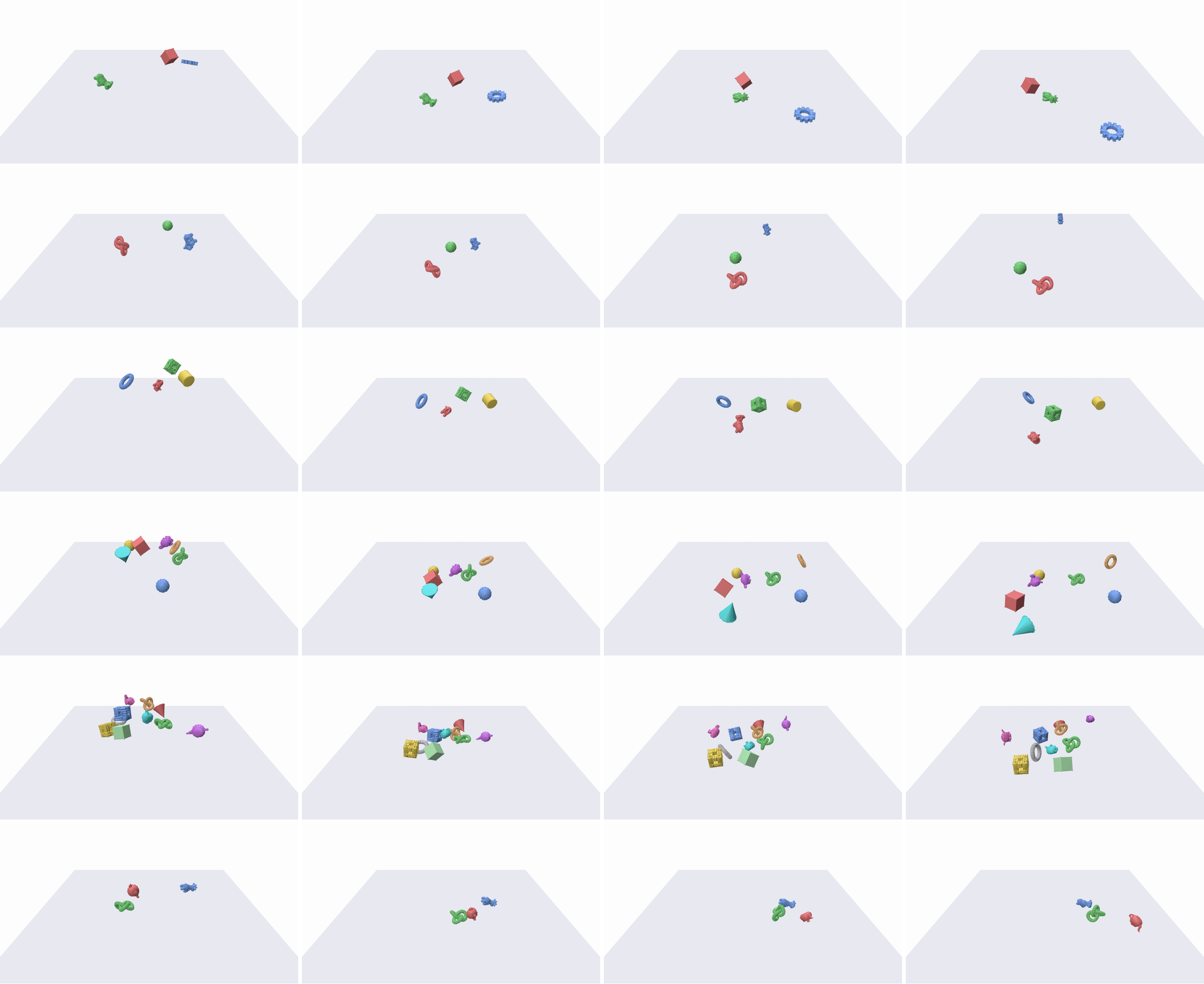}
    \vspace{-5mm}
\caption{\textbf{More qualitative results on MOVi-B.} Each row is one held-out test sample; columns show four frames from the rollout.}
    \label{supp_fig:more_vis_b}
\end{figure}

\begin{figure}
    \centering
    \includegraphics[width=\linewidth]{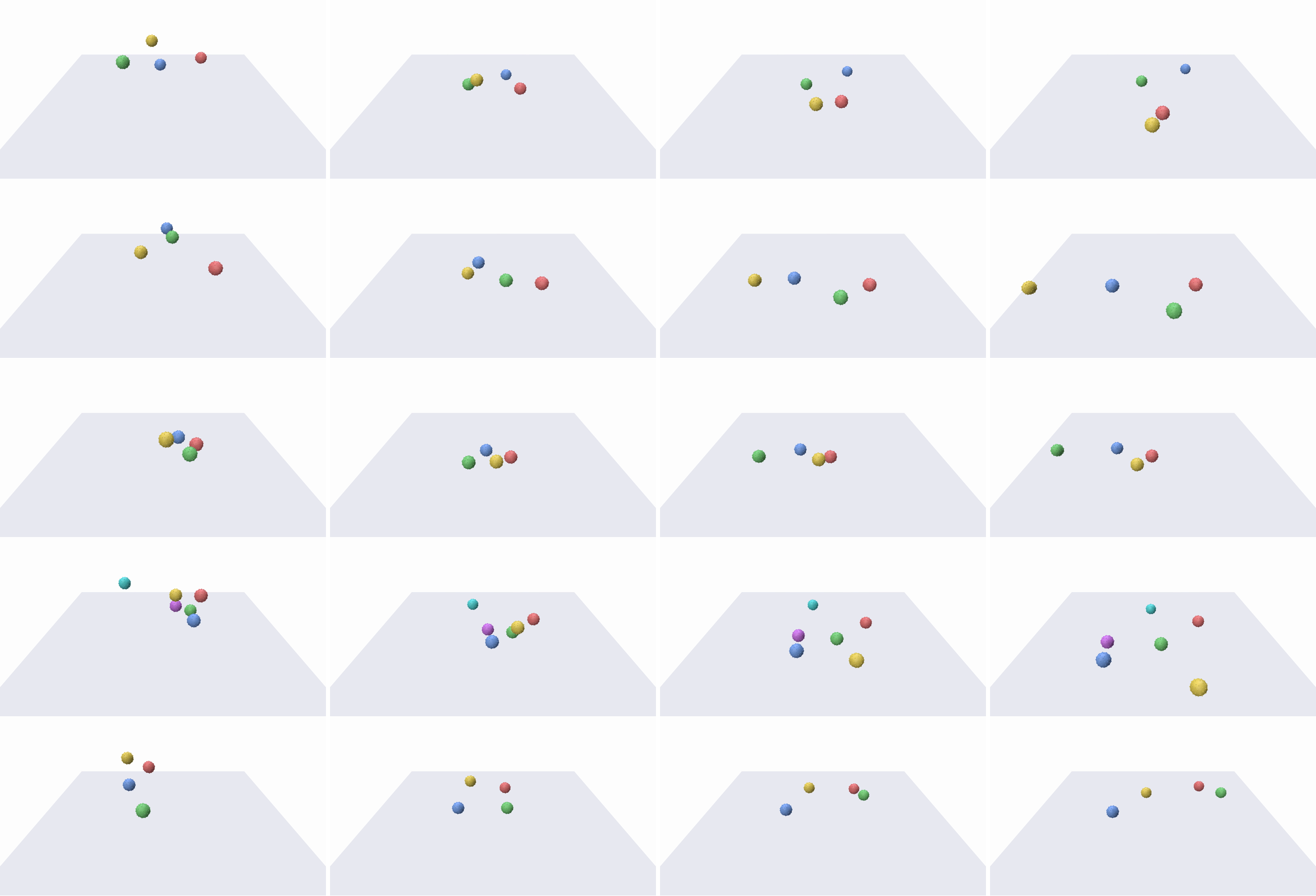}
    \vspace{-4mm}
\caption{\textbf{More qualitative results on MOVi-Sphere.} Each row is one held-out test sample; columns show four frames from the rollout.}
    \label{supp_fig:more_vis_s}
\end{figure}

We provide additional qualitative visualizations in Figs.~\ref{supp_fig:more_vis_a}--\ref{supp_fig:more_vis_s}, covering held-out MOVi-A, MOVi-B, and MOVi-Sphere test scenes. Each row corresponds to one selected sample and shows four uniformly sampled frames from the rollout. These examples complement Fig.~\ref{fig:main_visualization} and show that \name produces stable multi-object rollouts across different object geometries and contact configurations.

\section{Theoretical Properties of the Object-Anchor Representation}
\label{supp_sec:theoretical_properties}

We state the exact symmetry and structural guarantees used in Sec.~\ref{sec:anchor_rope}. These properties concern indexing choices and rigid projection in the representation; they do not claim full physical equivariance of the simulator under arbitrary rigid transformations of the world.

\paragraph{Anchor reindexing invariance.}
Let $\psi_\omega(\mathbf{x})$ denote the shared 3D rotary anchor map used by ARoPE, and define the object descriptor
\begin{equation}
\rho(A_i)=\frac{1}{N_a}\sum_{k=1}^{N_a}\psi_\omega(\mathbf{x}^{(i)}_k),
\qquad A_i=\{\mathbf{x}^{(i)}_k\}_{k=1}^{N_a}.
\end{equation}
For any permutation $\pi$ of the anchors,
\begin{equation}
\rho(\{\mathbf{x}^{(i)}_{\pi(k)}\}_{k=1}^{N_a})
=\frac{1}{N_a}\sum_{k=1}^{N_a}\psi_\omega(\mathbf{x}^{(i)}_{\pi(k)})
=\frac{1}{N_a}\sum_{k=1}^{N_a}\psi_\omega(\mathbf{x}^{(i)}_k)
=\rho(A_i).
\end{equation}
Thus the ARoPE descriptor is invariant to anchor ordering. The same statement holds for any shared per-anchor map followed by a symmetric pooling operator; we use mean pooling for simplicity and scale stability. Since $\psi_\omega$ is linear in the position, $\rho(A_i)=\psi_\omega(\bar{\mathbf{x}}^{(i)})$ with $\bar{\mathbf{x}}^{(i)}=\frac{1}{N_a}\sum_{k}\mathbf{x}^{(i)}_k$: the descriptor equals the rotary angles of the anchor centroid.

\paragraph{Vertex-order invariance of AVP.}
For a fixed anchor $\mathbf{q}^{(i,k)}$, AVP computes weights
\begin{equation}
w^{(i,k,v)}=\exp\!\left(-\frac{\|\mathbf{x}^{(i,v)}-\mathbf{q}^{(i,k)}\|}{\sigma}\right),
\qquad
\mathbf{u}^{(i,k)}=
\frac{\sum_v w^{(i,k,v)}\mathbf{f}^{(i,v)}}{\sum_v w^{(i,k,v)}} .
\end{equation}
If the vertices and their encoder features are reindexed by any permutation $\tau$, both the numerator and denominator are unchanged after relabeling the summation index. Therefore AVP is invariant to vertex ordering. The weights also depend only on Euclidean distances: for any common rigid transform $\mathbf{x}'=\mathbf{R}\mathbf{x}+\mathbf{t}$ and $\mathbf{q}'=\mathbf{R}\mathbf{q}+\mathbf{t}$ with $\mathbf{R}\in SO(3)$, $\|\mathbf{x}'-\mathbf{q}'\|=\|\mathbf{x}-\mathbf{q}\|$. Thus the AVP attention weights are unchanged by common rigid transforms of an object's current point and anchor coordinates.
In the batched implementation, weights on padded vertices are set to zero before normalization, which is equivalent to summing only over valid vertices.

\paragraph{Object permutation equivariance.}
Consider a permutation matrix $P$ acting on the $M$ object tokens, and let $\tilde P=\mathrm{diag}(P,I_{N_r})$ act on the concatenation of object tokens and the $N_r{=}16$ register tokens. Self-attention without sequence-index positional embeddings satisfies
\begin{equation}
\mathrm{Attn}(\tilde P\mathbf{Q},\tilde P\mathbf{K},\tilde P\mathbf{V})
=\tilde P\,\mathrm{Attn}(\mathbf{Q},\mathbf{K},\mathbf{V}),
\end{equation}
because the row-wise softmax and value aggregation commute with simultaneous token permutation. Token-wise RMSNorm, FiLM, elementwise gated attention, and FFNs also commute with $\tilde P$ since their parameters are shared across object positions. Therefore each decoder block maps a permutation of object tokens to the same permutation of outputs, while the register tokens remain a shared permutation-invariant workspace. Applying the same argument layer by layer proves equivariance of the object-level decoder.

\paragraph{Anchor prediction and rigid projection.}
The anchor predictor uses shared weights for each object and applies cross-object attention to the permuted object-token set; hence permuting object order only permutes the predicted anchor updates by object. Within one object, Kabsch alignment depends on sums of corresponding centered source and target anchors. If source and target anchors are reindexed by the same permutation, the cross-covariance matrix is unchanged, so the recovered rigid transform $(\mathbf{R}^{(i)},\mathbf{t}^{(i)})$ is unchanged. The full simulator is therefore equivariant to object reordering and invariant to anchor reindexing under consistent anchor correspondences.

\paragraph{Rigidity by construction.}
After Kabsch alignment, every output vertex is generated as
\begin{equation}
\mathbf{x}_{t+1}^{(i,v)}=\mathbf{R}^{(i)}\mathbf{x}_{\mathrm{ref}}^{(i,v)}+\mathbf{t}^{(i)},
\qquad \mathbf{R}^{(i)}\in SO(3).
\end{equation}
For any two vertices $u,v$ of the same object,
\begin{equation}
\|\mathbf{x}_{t+1}^{(i,u)}-\mathbf{x}_{t+1}^{(i,v)}\|_2
=\|\mathbf{R}^{(i)}(\mathbf{x}_{\mathrm{ref}}^{(i,u)}-\mathbf{x}_{\mathrm{ref}}^{(i,v)})\|_2
=\|\mathbf{x}_{\mathrm{ref}}^{(i,u)}-\mathbf{x}_{\mathrm{ref}}^{(i,v)}\|_2 .
\end{equation}
Thus intra-object distances are preserved exactly by the final projection, independent of the raw anchor accelerations predicted by the network. The simulator is intentionally not SE(3)-equivariant overall: gravity, ground contact, and evaluation coordinates are expressed in the world frame.

\section{Training Details}
\label{supp_sec:training}
\label{supp_section:training_details}

\subsection{Training Schedule}
The final model is trained directly with sequence length $T{=}8$ and random integration step sizes $\Delta t\in\{1,5,10\}$ sampled with near-uniform probabilities. The same FiLM-conditioned model is therefore exposed to short and long temporal discretizations throughout training. Note that we did not use curriculum learning or scheduled sampling in the reported MOVi experiments.

We optimize with AdamW~\cite{loshchilov2019decoupled} using a base learning rate of $10^{-4}$, weight decay $0.01$, $\beta_1{=}0.9$, and $\beta_2{=}0.999$. The learning rate follows a 10-epoch linear warmup from 10\% of the base rate, followed by cosine decay to $10^{-6}$. Gradients are clipped to norm $1.0$. Training runs for 300 epochs with 126 (MOVi-B) or 128 (MOVi-A, MOVi-Sphere) training sequences per step. We use HopNet-style $960/120/120$ train/validation/test split files for each MOVi variant. For a fair comparison, we follow the evaluation of HopNet~\cite{wei2025integrating}, which gives four ground-truth frames (frames 0--3) followed by an autoregressive rollout; at larger step sizes, the given frames are spaced by the step.

\noindent\textbf{VPD.} Our VPD baseline uses 512 points per object; a scene with more than 1{,}536 points is reduced to a fixed random subset of 1{,}536 points, so that training fits in the memory of one GPU.

\subsection{Loss Implementation Details}
\label{supp_sec:loss_details}

All loss terms in Sec.~\ref{sec:training_obj} are implemented using the Smooth L1 Loss~\cite{girshick2015fast}~(PyTorch \texttt{nn.SmoothL1Loss()} function\footnote{\url{https://docs.pytorch.org/docs/stable/generated/torch.nn.SmoothL1Loss.html}}), applied elementwise at anchor locations; the implementation sums coordinate losses over $x,y,z$ and averages over valid anchors, objects, and time steps. Concretely, for two tensors $\mathbf{u}$ and $\mathbf{v}$ of the same shape, the Smooth L1 loss is defined as
\begin{equation}
\ell_{\text{SL1}}(\mathbf{u}, \mathbf{v})
=
\frac{1}{|\mathbf{u}|}
\sum_j
\begin{cases}
\frac{1}{2}(u_j - v_j)^2, & \text{if } |u_j - v_j| < 1,\\[4pt]
|u_j - v_j| - \frac{1}{2}, & \text{otherwise},
\end{cases}
\end{equation}
which corresponds to the default setting of \texttt{nn.SmoothL1Loss()} in PyTorch. Compared with $\ell_2$ loss, Smooth L1 is less sensitive to occasional large errors during autoregressive rollout, while remaining smooth near zero and thus stable for optimization.

In our training objective, both the position and acceleration losses are computed using Smooth L1 before and after rigid projection. All four terms are evaluated at anchor locations. Specifically, the raw position loss is
\begin{equation}
\mathcal{L}_{\text{pos}}^{\text{raw}}
=
\ell_{\text{SL1}}\!\left(
    \hat{\mathbf{q}}_{t+1}^{\text{raw}},
    \mathbf{q}_{t+1}^{\text{gt}}
\right),
\end{equation}
and the rigid position loss is
\begin{equation}
\mathcal{L}_{\text{pos}}^{\text{rigid}}
=
\ell_{\text{SL1}}\!\left(
    \hat{\mathbf{q}}_{t+1}^{\text{rigid}},
    \mathbf{q}_{t+1}^{\text{gt}}
\right),
\end{equation}
where $\hat{\mathbf{q}}_{t+1}^{\text{raw}}$ denotes the predicted anchor positions before Kabsch alignment, $\hat{\mathbf{q}}_{t+1}^{\text{rigid}}$ denotes the corresponding anchors after rigid projection, and $\mathbf{q}_{t+1}^{\text{gt}}$ is the ground-truth anchor state. Full-resolution vertices are supervised indirectly through the rigid transform induced by the anchor update.

Similarly, the raw and rigid acceleration losses are defined as
\begin{equation}
\mathcal{L}_{\text{acc}}^{\text{raw}}
=
\ell_{\text{SL1}}\!\left(
\hat{\mathbf{a}}_{t}^{\text{raw}},
\mathbf{a}_{t}^{\text{gt}}
\right),
\qquad
\mathcal{L}_{\text{acc}}^{\text{rigid}}
=
\ell_{\text{SL1}}\!\left(
\hat{\mathbf{a}}_{t}^{\text{rigid}},
\mathbf{a}_{t}^{\text{gt}}
\right),
\end{equation}
where the ground-truth acceleration is computed from the trajectory using the same Verlet discretization,
\begin{equation}
\mathbf{a}_{t}^{\text{gt}}
=
\frac{\mathbf{q}_{t+1}^{\text{gt}} - 2\mathbf{q}_{t}^{\text{gt}} + \mathbf{q}_{t-1}^{\text{gt}}}{\Delta t^2}.
\end{equation}

The final loss is therefore
\begin{equation}
\mathcal{L}
=
\lambda_{\text{pos}}
\left(
\mathcal{L}_{\text{pos}}^{\text{raw}}
+
\mathcal{L}_{\text{pos}}^{\text{rigid}}
\right)
+
\lambda_{\text{acc}}
\left(
\mathcal{L}_{\text{acc}}^{\text{raw}}
+
\mathcal{L}_{\text{acc}}^{\text{rigid}}
\right),
\end{equation}
with $\lambda_{\text{pos}}=10$ and $\lambda_{\text{acc}}=1$ in the reported experiments. For multi-step training, residuals are normalized by $\Delta t^2$ before the Smooth L1 reduction, matching the acceleration scale used by Verlet integration.

\section{Data Statistics}
\label{supp_section:data_statistics}

As detailed in Tab.~\ref{tab:movi_objects}, the datasets vary in
object complexity. \textbf{MOVi-A} features simple geometric primitives---cubes
(51 vertices), cylinders (64 vertices), and spheres (64 vertices)---uniformly
distributed across scenes. \textbf{MOVi-B} introduces 11 object categories with
diverse mesh complexities ranging from 51 vertices (cube) to 1{,}142 vertices
(torus\_knot), including complex shapes such as gear (569 vertices), sponge
(908 vertices), spot (682 vertices), and suzanne (523 vertices); all categories
appear with approximately equal frequency ($\sim$9\% each).
\textbf{MOVi-Sphere} serves as a controlled baseline containing only spheres
(64 vertices each). This diversity enables comprehensive evaluation across
varying geometric complexities, from simple 51-vertex primitives to intricate
1{,}142-vertex meshes. Each of MOVi-A, MOVi-B and MOVi-Sphere contains 1{,}200 scenes simulated for
480 frames. \textbf{MOVi-C}, used only for the zero-shot evaluation (Tab.~\ref{tab:movic}, Appendix~\ref{supp_sec:movic}),
consists of the 120 test scenes provided by HopNet~\cite{wei2025integrating}, each with 3 to 10 scanned everyday
objects from Google Scanned Objects~\cite{downs2022google}: 768 objects of 518 distinct assets with 46 to 1{,}908
vertices (mean 540), simulated for 480 frames.

\begin{table}[h]
  \centering
  \caption{\textbf{Object Types and Vertex Counts in MOVi Benchmarks.}}
  \label{tab:movi_objects}
  \vspace{1mm}
  \small
  \begin{tabular}{llcc}
  \toprule
  \textbf{Dataset} & \textbf{Object Name} & \textbf{Vertices} & \textbf{Ratio} \\
  \midrule
  \multirow{3}{*}{MOVi-A}
   & cube & 51 & 33.6\% \\
   & cylinder & 64 & 33.8\% \\
   & sphere & 64 & 32.6\% \\
  \midrule
  \multirow{11}{*}{MOVi-B}
   & cone & 64 & 8.8\% \\
   & cube & 51 & 9.3\% \\
   & cylinder & 64 & 8.8\% \\
   & gear & 569 & 8.8\% \\
   & sphere & 64 & 9.2\% \\
   & sponge & 908 & 8.7\% \\
   & spot & 682 & 9.7\% \\
   & suzanne & 523 & 9.1\% \\
   & teapot & 274 & 9.4\% \\
   & torus & 647 & 8.9\% \\
   & torus\_knot & 1142 & 9.2\% \\
  \midrule
  \multirow{1}{*}{MOVi-Sphere}
   & sphere & 64 & 100.0\% \\
  \midrule
  MOVi-C & 518 scanned objects & 46--1{,}908 & $\le$0.7\% each \\
  \bottomrule
  \end{tabular}
\end{table}

\begin{table}[h]
  \centering
  \caption{\textbf{Dataset summary statistics}.}
  \label{tab:movi_summary}
  \vspace{1mm}
  \small
  \begin{tabular}{lcccc}
  \toprule
  \textbf{Dataset} & \textbf{Scenes} & \textbf{Total instances} & \textbf{Object types} & \textbf{Frames/scene} \\
  \midrule
  MOVi-A      & 1{,}200 & 7{,}810 & 3 primitives  & 480 \\
  MOVi-B      & 1{,}200 & 7{,}764 & 11 shapes     & 480 \\
  MOVi-Sphere & 1{,}200 & 7{,}810 & 1 primitive   & 480 \\
  MOVi-C      & 120 (test) & 768 & 518 scanned objects & 480 \\
  \bottomrule
  \end{tabular}
\end{table}

\subsection{Cross-Dataset Transfer Analysis}
\label{supp_sec:cross_dataset_analysis}

Tab.~\ref{tab:generalization} shows that the MOVi-Sphere-trained model transfers least well to the other MOVi variants. Averaged over the two held-out target datasets, the 100-step position errors of the MOVi-Sphere-trained model are $0.172$, $0.151$, and $0.117$ for step sizes $1$, $5$, and $10$, respectively. The corresponding averages are $0.137$, $0.115$, and $0.095$ for MOVi-B training, and $0.116$, $0.095$, and $0.080$ for MOVi-A training. The gap is larger in translation than in orientation: from MOVi-Sphere to MOVi-A, the orientation error at 100 steps is close to that of MOVi-B training ($17.57^\circ$ against $17.18^\circ$), while the position error is $24\%$ higher ($0.205$ against $0.165$). One interpretation is related to geometry. MOVi-Sphere contains only spheres, whose contacts involve no edges or faces, whereas MOVi-A contains three primitives and MOVi-B 11 categories with much broader shape variation (Tab.~\ref{tab:movi_objects}); a model trained on spheres sees fewer of the contact configurations that occur in the other datasets. This interpretation is meant only to explain the average trend; it is not evidence that cross-dataset transfer is generally easier than in-domain evaluation.

\section{Data Augmentation}
\label{supp_sec:augmentation}

To improve generalization and prevent overfitting, we employ two complementary data augmentation strategies during training:

\paragraph{Rotation Augmentation.}
We apply random Z-axis rotations to the entire scene during training. Given the original vertex positions $\mathbf{X} \in \mathbb{R}^{N \times 3}$, we apply:
\begin{equation}
\mathbf{X}' = \mathbf{X} \mathbf{R}_z(\theta)^\top
\end{equation}
where $\mathbf{R}_z(\theta)$ is a rotation matrix around the Z-axis (gravity direction). The rotation angle $\theta$ is sampled from discrete values $\{5^\circ, 10^\circ, 15^\circ, \ldots, 355^\circ\}$ with angle step $5^\circ$. Rotation is sampled once per training batch: with probability $0.5$, the whole batch is rotated by the same sampled angle. This improves robustness to yaw rotations around the gravity axis rather than imposing full 3D rotation invariance, which would be inappropriate for scenes with gravity and ground contact.

Importantly, the rotation is applied consistently to all objects in a scene and to all frames in a sequence, preserving the relative object relationships and temporal dynamics. Since the full trajectory is rotated before acceleration targets are computed, accelerations transform consistently:
\begin{equation}
\mathbf{a}' = \mathbf{R}_z(\theta) \mathbf{a}
\end{equation}

\paragraph{Object Permutation Augmentation.}
Multi-object dynamics should be permutation-equivariant: the physics outcome should not depend on the arbitrary ordering of objects in the input. The architecture already satisfies this property for object tokens (Appendix~\ref{supp_sec:theoretical_properties}); during training, we additionally randomly permute object ordering with 50\% probability as a robustness check. Given $N_{\text{obj}}$ objects, we sample a random permutation $\pi \in S_{N_{\text{obj}}}$ and reorder all per-object tensors accordingly:
\begin{equation}
\{\mathbf{X}^{(1)}, \ldots, \mathbf{X}^{(N_{\text{obj}})}\} \rightarrow \{\mathbf{X}^{(\pi(1))}, \ldots, \mathbf{X}^{(\pi(N_{\text{obj}}))}\}
\end{equation}

This augmentation is applied in the data loader before collation and feature computation, ensuring consistent permutation across all object attributes (positions, velocities, physics parameters). It helps detect implementation mistakes that would otherwise introduce order dependence and improves robustness to arbitrary input arrangements at inference time.

Rotation is applied after moving each training batch to the GPU, while object permutation is applied on the data-loading side before collation. Both add negligible overhead to the training pipeline.

\section{Runtime Performance and Computational Costs}
\subsection{Runtime Performance}
\label{supp_sec:runtime}

We profile inference on MOVi-B scene 72, containing $10$ objects with a total of $4{,}016$ vertices, using a $50$-step rollout setting. Per-object vertex counts are $\{64, 51, 1142, 64, 64, 682, 682, 647, 51, 569\}$, covering diverse MOVi-B meshes including torus\_knot ($1{,}142$), spot ($682$), torus ($647$), gear ($569$), cube ($51$), and several primitives ($64$ vertices each). As summarized in Tab.~\ref{tab:inference_breakdown}, the model uses $107.0$M parameters at inference ($148.4$M including disabled optional modules) and has a peak memory footprint of $0.82$~GB.

We time each part with CUDA events at its boundaries; the events add about $5\%$ to the step, so the parts sum to $12.43$~ms instead of the $11.96$~ms of Tab.~\ref{tab:runtime_comparison}. The model core (point encoder + object-level interaction + anchor-object interaction) takes $9.99$~ms per step ($80.3\%$). Within the model core, the point encoder takes $1.78$~ms ($14.3\%$), object-level interaction takes $2.46$~ms ($19.8\%$), and anchor-object interaction takes $5.75$~ms ($46.2\%$). Geometric processing takes $1.12$~ms per step ($9.0\%$): the CUDA KNN search $0.43$~ms ($3.5\%$) and the differentiable Kabsch transform $0.69$~ms ($5.6\%$). The remaining $1.32$~ms ($10.6\%$) covers integration, ground-plane handling, anchor-vertex pooling and bookkeeping. These results suggest that further speedups should primarily target the anchor-object interaction.

\begin{table}[h]
\centering
\caption{\name inference performance on the MOVi-B evaluation set (times with CUDA events at the boundaries of each part; $11.96$~ms per step, $83.6$ FPS, without them).}
\label{tab:inference_breakdown}
\begin{tabular}{lcc}
\toprule
\multicolumn{3}{c}{\textbf{Overall Performance}} \\
\midrule
Parameters (inference / total) & \multicolumn{2}{c}{107.0M / 148.4M} \\
Peak Memory & \multicolumn{2}{c}{0.82 GB} \\
Total Objects & \multicolumn{2}{c}{10} \\
Total Vertices & \multicolumn{2}{c}{4016} \\
\midrule
\multicolumn{3}{c}{\textbf{Model Core Components}} \\
\midrule
Component & Time (ms) & Percent \\
\midrule
Point Encoder & 1.78 & 14.3\% \\
Object-Level Interaction & 2.46 & 19.8\% \\
Anchor-Object Interaction & 5.75 & 46.2\% \\
\textbf{Subtotal} & \textbf{9.99} & \textbf{80.3\%} \\
\midrule
\multicolumn{3}{c}{\textbf{Geometric Processing}} \\
\midrule
CUDA KNN & 0.43 & 3.5\% \\
Diff Kabsch Transform & 0.69 & 5.6\% \\
\textbf{Subtotal} & \textbf{1.12} & \textbf{9.0\%} \\
\midrule
Other (integration, ground plane, pooling, bookkeeping) & 1.32 & 10.6\% \\
\midrule
\textbf{Total} & \textbf{12.43} & \textbf{100\%} \\
\bottomrule
\end{tabular}
\end{table}

\paragraph{Vertex-Level Processing Analysis.}
To illustrate the efficiency of our object-level tokenization, we analyze the hypothetical computational cost if attention were computed at the vertex level. With $N_v = 4{,}016$ vertices, a single self-attention layer would require $4 \times N_v^2 \times D = 4 \times 4016^2 \times 768 \approx 49.5$ GFLOPs. In contrast, our object-level decoder self-attends over 10 object tokens and 16 unpositioned register tokens, requiring $4 \times 26^2 \times 768 \approx 2.1$ MFLOPs per layer for the attention matrix-value product. This is a ${\sim}2.4{\times}10^4$ reduction in the quadratic attention term before accounting for linear projections and FFNs. The measured runtime in Tab.~\ref{tab:inference_breakdown} includes the full model and geometric processing; the main cost is the anchor-object interaction rather than object-level attention.

\section{Detailed Network Architecture}
\label{supp_sec:architecture}

Tab.~\ref{tab:arch_overview} summarizes the architecture used in the main experiments. The model uses 107.0M parameters at inference and consists of a hierarchical PointNet encoder, a 4-layer object-level Transformer decoder with gated attention and register tokens, and a query-based anchor predictor. The publicly available weights additionally contain 41.4M parameters of optional modules that are disabled in all reported experiments (148.4M in total).

\begin{table}[h]
  \centering
  \caption{\textbf{Architecture Overview.} Summary of model components and their configurations.}
  \label{tab:arch_overview}
  \vspace{1mm}
  \resizebox{0.78\columnwidth}{!}{%
  \begin{tabular}{llc}
  \toprule
  \textbf{Component} & \textbf{Configuration} & \textbf{Notes} \\
  \midrule
  \multicolumn{3}{l}{\textit{Input Configuration}} \\
  \quad Max objects per scene & $M = 16$ & -- \\
  \quad Vertices per object & all, no limit & up to 1{,}142 in training, 1{,}908 in MOVi-C \\
  \quad FPS anchors per object & $N_a = 4$ & -- \\
  \quad Encoder input dim & 12D & -- \\
  \quad \quad -- Displacement vector & $\mathbf{d} \in \mathbb{R}^3$ & nearest contact geometry \\
  \quad \quad -- Velocity & $\mathbf{v} \in \mathbb{R}^3$ & $\mathbf{x}_t - \mathbf{x}_{t-1}$ \\
  \quad \quad -- Relative position & $\mathbf{r} \in \mathbb{R}^3$ & $\mathbf{x}_t - \mathbf{x}_{\text{ref}}$ \\
  \quad \quad -- Physics params & $[m,\mu,\epsilon] \in \mathbb{R}^3$ & mass, friction, restitution \\
  \quad Time step & $[s, s^2] \in \mathbb{R}^2$ & per-layer FiLM conditioning \\
  \midrule
  \multicolumn{3}{l}{\textit{Encoder (Hierarchical PointNet)}} \\
  \quad Backbone convolutions & Conv1d MLP & shared over objects \\
  \quad Hierarchical pooling & 4 levels (100\%, 50\%, 25\%, 12.5\%) & multi-scale geometry \\
  \quad Object embedding & $D=768$ & fixed-size object token \\
  \midrule
  \multicolumn{3}{l}{\textit{Decoder (4-Layer Transformer)}} \\
  \quad Hidden dimension & $D = 768$ & -- \\
  \quad Attention heads & $H = 6$ (head dim = 128) & -- \\
  \quad Object self-attention & Gated SDPA & elementwise sigmoid gate \\
  \quad Register tokens & 16 learnable tokens & unpositioned global workspace \\
  \quad Time conditioning & per-layer FiLM & code $[s,s^2]$ \\
  \quad FFN & SwiGLU, $2.5\times$ expansion & RMSNorm + QK norm \\
  \quad Block attention residuals & block size $S{=}4$ & inter-block residual~\cite{chen2026attnres} \\
  \quad Dropout & $0.1$ & attention and FFN \\
  \midrule
  \multicolumn{3}{l}{\textit{Anchor Predictor}} \\
  \quad Anchor queries & $N_a$ per object & FPS anchors \\
  \quad Anchor-Vertex Pooling & 256D feature & learned isotropic distance kernel \\
  \quad Predictor input & 271D & anchor state + AVP/context features \\
  \quad Cross-attention & object and cross-object context & anchor-level prediction \\
  \quad Output & 3D anchor acceleration & followed by Verlet + Kabsch \\
  \midrule
  \multicolumn{3}{l}{\textit{Position Encoding and Projection}} \\
  \quad ARoPE dimension & 96 & set-aggregated anchor RoPE \\
  \quad Rigid projection & Kabsch via RoMa & differentiable alignment \\
  \midrule
  \textbf{Total Parameters} & 107.0M at inference & 148.4M incl.\ disabled modules \\
  \bottomrule
  \end{tabular}
  }
\end{table}

\section{Zero-Shot Transfer to MOVi-C}
\label{supp_sec:movic}

\noindent\textbf{Data.} MOVi-C~\cite{greff2022kubric} scenes hold 3 to 10 objects from Google Scanned Objects~\cite{downs2022google}. We use the 120 test scenes provided by HopNet~\cite{wei2025integrating} (480 frames at 240 frames per second; 768 objects of 518 assets) in the format of our MOVi-A, MOVi-B and MOVi-Sphere data: each object is represented by the vertices of its collision mesh (46 to 1{,}908 vertices, mean 540, without decimation), posed by the recorded positions and orientations. Kubric scales an asset by the recorded scale divided by the largest extent of the asset's bounding box, so that its largest dimension equals the recorded scale; with this factor (1.9 to 58.9), the meshes reproduce the recorded 3D bounding boxes to a relative error of $10^{-7}$, and every resting object lies 2\,mm above the floor, as in our other data. The provided metadata list the objects that are visible in the video; in 10 scenes, 11 simulated objects are never visible and are therefore not included. We use the scenes as provided, as HopNet does.

\noindent\textbf{Physical parameters.} The model reads the recorded masses, friction and restitution. The masses equal the volumes of the unscaled assets ($10^{-5}$ to 0.036) and lie below every training mass (at least 0.1); friction and restitution (0.5 for every object) lie between the training values; and 313 of the 768 objects are larger than any training object.

\noindent\textbf{Protocol.} The models of Tab.~\ref{tab:main_comparison} are applied without any change and evaluated as in Sec.~\ref{sec:exp}: four given frames (frames 0 to 3 at step size 1, spaced by the step otherwise), position and orientation errors per object, averaged over the 120 scenes. HopNet's values are those reported in~\cite{wei2025integrating} (zero-shot, 50 frames). Tab.~\ref{supp_tab:movic} reports our errors up to 100 frames.

\begin{table}[h]
\centering
\caption{\textbf{Zero-shot transfer to MOVi-C}: \name trained on each MOVi dataset, at step sizes 1, 5 and 10. Each cell reports position RMSE (m)\,/\,orientation RMSE ($^{\circ}$); step size 10 predicts only frames that are multiples of 10 (--).}
\label{supp_tab:movic}
\vspace{1mm}
\small
\begin{tabular}{llcccc}
\toprule
& & \multicolumn{4}{c}{\textbf{Rollout horizon (frames)}} \\
\cmidrule(lr){3-6}
\textbf{Trained on} & \textbf{Step size} & 25 & 50 & 75 & 100 \\
\midrule
\multirow{3}{*}{MOVi-Sphere} & 1 & 0.026\,/\,2.02 & 0.096\,/\,7.55 & 0.212\,/\,15.68 & 0.367\,/\,25.58 \\
 & 5 & 0.008\,/\,0.62 & 0.061\,/\,4.34 & 0.160\,/\,10.82 & 0.298\,/\,19.13 \\
 & 10 & -- & 0.032\,/\,2.26 & -- & 0.240\,/\,15.78 \\
\midrule
\multirow{3}{*}{MOVi-A} & 1 & 0.026\,/\,1.93 & 0.095\,/\,6.95 & 0.204\,/\,14.61 & 0.350\,/\,24.64 \\
 & 5 & 0.008\,/\,0.58 & 0.058\,/\,4.01 & 0.149\,/\,9.97 & 0.272\,/\,17.77 \\
 & 10 & -- & 0.032\,/\,2.12 & -- & 0.228\,/\,15.28 \\
\midrule
\multirow{3}{*}{MOVi-B} & 1 & 0.021\,/\,1.60 & 0.086\,/\,5.88 & 0.187\,/\,12.25 & 0.320\,/\,20.38 \\
 & 5 & 0.007\,/\,0.51 & 0.055\,/\,3.56 & 0.147\,/\,9.13 & 0.272\,/\,16.58 \\
 & 10 & -- & 0.032\,/\,2.02 & -- & 0.229\,/\,14.63 \\
\bottomrule
\end{tabular}
\end{table}

\section{Partial Point Clouds from a Single Depth Camera}
\label{supp_sec:partial_depth}

\noindent\textbf{Inputs.} A pinhole depth camera ($640\times480$ pixels, $58^\circ$ vertical field of view) observes each test scene once, at frame 3, from 12\,m along the viewing direction of the scene's camera, aimed at a point 1.5\,m above the center of the arena. Each object is represented by the back-projected points of its segment in its own frame, subsampled at random to at most 2{,}048 points and then to at most 1{,}024 by farthest point sampling (FPS). Objects with fewer than 16 points are left out.

\noindent\textbf{Completion network.} A PointNet encoder with a multilayer perceptron decoder takes the first 256 observed points of an object, the camera elevation and distance, and the height of the points above the ground, and predicts 512 points on the object's surface; FPS reduces the observed and predicted points to at most 1{,}024 (40--45\% predicted). It is trained on the 960 MOVi-B training scenes and 2{,}400 further scenes from the MOVi-B shape library, which includes the shapes of MOVi-A and MOVi-Sphere, and then kept fixed.

\noindent\textbf{Model.} The architecture is unchanged: the observed points, or the observed and completed points, in FPS order, take the place of the mesh vertices, and the first four are the anchors. The model is given frames 0 to 3, with the points of frame 3 placed in frames 0 to 3 by the ground-truth poses; the camera is not used afterwards. The rigid motion of each object moves all of its points and, for evaluation, its collision mesh, which is scored as in the main protocol over the observed objects.

\noindent\textbf{Training.} Starting from the weights in Tab.~\ref{tab:main_comparison}, we train the model of each dataset on partial point clouds with one recipe: each sample is observed once by a camera like the evaluation camera at a random azimuth and completed by the fixed network, and the losses use the ground-truth positions of all points. We update all weights and the batch-normalization statistics with AdamW ($\beta_1=0.9$, $\beta_2=0.999$, weight decay 0.01 on weight matrices), a learning rate of $4\times10^{-6}$ ($10\times$ for the fusion layer of the acceleration head), a 2-epoch linear warmup from $0.1\times$, a cosine decay to $5\times10^{-8}$ over 60 epochs, gradient clipping at 1.0 and 21 scenes per step, and evaluate the exponential moving average of the weights (decay 0.9999). Without completion, we train only the encoder and the decoder, keeping the acceleration head and the batch-normalization statistics of Tab.~\ref{tab:main_comparison}, with the same schedule and an additional linear warmup from zero over the first epoch.

\noindent\textbf{Visualization.} Fig.~\ref{fig:partial_examples} shows frames 240 and 476 (1\,s and 1.98\,s at 240 frames per second, step size 1). Observed points are opaque in their object's color, completed points smaller and lighter, and the mesh, moved by the same rigid motion, translucent. The depth and segmentation images are $640\times360$ center crops of the inputs.

\begin{table}[h]
  \centering
  \caption{\textbf{Partial point clouds} (test scenes, step size 1): position RMSE (m)\,/\,orientation RMSE ($^{\circ}$) at 50, 75 and 100 frames. Full: the mesh vertices, as in Tab.~\ref{tab:main_comparison}. Partial: the observed points of each object only. Partial + completion: the observed points with those of the completion network. Tab.~\ref{tab:main_comparison}: the models of Tab.~\ref{tab:main_comparison}, trained on full point clouds and applied unchanged. Trained: these models trained further on the given input (Training above).}
  \label{supp_tab:partial}
  \vspace{1mm}
  \resizebox{\textwidth}{!}{%
  \begin{tabular}{llccccccccc}
  \toprule
  & & \multicolumn{3}{c}{\textbf{MOVi-A}} & \multicolumn{3}{c}{\textbf{MOVi-B}} & \multicolumn{3}{c}{\textbf{MOVi-Sphere}} \\
  \cmidrule(lr){3-5}\cmidrule(lr){6-8}\cmidrule(lr){9-11}
  \textbf{Input} & \textbf{Model} & 50 & 75 & 100 & 50 & 75 & 100 & 50 & 75 & 100 \\
  \midrule
  Full & Tab.~\ref{tab:main_comparison} & 0.040\,/\,4.81 & 0.086\,/\,10.64 & 0.145\,/\,17.57 & 0.030\,/\,3.29 & 0.061\,/\,7.16 & 0.108\,/\,12.65 & 0.022\,/\,2.93 & 0.045\,/\,6.28 & 0.078\,/\,10.74 \\
  \midrule
  Partial & Tab.~\ref{tab:main_comparison} & 0.055\,/\,4.84 & 0.111\,/\,10.69 & 0.182\,/\,17.84 & 0.041\,/\,3.75 & 0.083\,/\,8.04 & 0.142\,/\,14.29 & 0.053\,/\,3.30 & 0.110\,/\,7.15 & 0.184\,/\,12.65 \\
  Partial & trained & 0.055\,/\,4.82 & 0.110\,/\,10.52 & 0.182\,/\,17.53 & 0.040\,/\,3.67 & 0.080\,/\,7.90 & 0.136\,/\,13.90 & 0.053\,/\,3.26 & 0.108\,/\,7.11 & 0.180\,/\,12.60 \\
  \midrule
  Partial + completion & Tab.~\ref{tab:main_comparison} & 0.052\,/\,5.24 & 0.097\,/\,10.73 & 0.162\,/\,18.02 & 0.036\,/\,3.75 & 0.075\,/\,7.88 & 0.129\,/\,13.48 & 0.023\,/\,3.08 & 0.048\,/\,6.54 & 0.083\,/\,11.10 \\
  Partial + completion & trained & 0.047\,/\,4.71 & 0.092\,/\,9.78 & 0.156\,/\,16.44 & 0.034\,/\,3.54 & 0.069\,/\,7.54 & 0.120\,/\,13.21 & 0.026\,/\,2.92 & 0.054\,/\,6.23 & 0.093\,/\,10.63 \\
  \bottomrule
  \end{tabular}}
\end{table}

\section{Large-Scale Simulation}
\label{supp_sec:large_scale}

To evaluate the scalability of \name to scenes with significantly more objects than the MOVi benchmarks, we construct the WreckingBall dataset featuring dense multi-object collision scenarios.

\subsection{Dataset Generation}
\label{supp_sec:wb_data_synthesis}

We simulate a wrecking ball scenario where a spherical projectile collides with a wall of stacked cubes. Four scene configurations are generated with varying grid sizes:
\begin{itemize}
    \item \textbf{WreckingBall-28}: $3 \times 3 \times 3$ cube arrangement ($M{=}27$ cubes + 1 ball = 28 objects)
    \item \textbf{WreckingBall-65}: $4 \times 4 \times 4$ cube arrangement ($M{=}64$ cubes + 1 ball = 65 objects)
    \item \textbf{WreckingBall-126}: $5 \times 5 \times 5$ cube arrangement ($M{=}125$ cubes + 1 ball = 126 objects)
    \item \textbf{WreckingBall-217}: $6 \times 6 \times 6$ cube arrangement ($M{=}216$ cubes + 1 ball = 217 objects)
\end{itemize}

\paragraph{Data Generation Parameters.}
For each grid configuration, we generate 20 independent simulation samples, yielding a total of 80 samples across all configurations. The dataset is split into 72 training samples (18 per grid size) and 8 test samples (2 per grid size). Each trajectory spans $T{=}600$ frames at native simulation frequency. Cubes are placed in a deterministic regular grid with $1.05$\,m spacing. Per-sample variation comes from a randomized vertical offset of the ball's launch position (sampled uniformly from $[2.5, 4.0]$\,m); the launch velocity is fixed at $30$\,m/s along the $+x$ axis toward the cube wall.

\paragraph{Reading the Four Configurations.}
The configurations differ in more than the number of cubes. The projectile has the same size and launch speed in all of them, while the wall grows from $3{\times}3{\times}3$ to $6{\times}6{\times}6$ cubes; the size of the projectile relative to the wall, the height of impact relative to the wall, the energy delivered per cube and the density of contacts therefore change together with the cube count. The four configurations are thus stress tests of the same model, not a controlled ablation of the object count.

\paragraph{Output Data Format.}
Each sample contains the following per-frame data:
\begin{itemize}
    \item \textbf{Vertex positions}: Per-object vertex positions $\mathbf{X}^{(i)}_t \in \mathbb{R}^{N_v^{(i)} \times 3}$ for each frame.
    \item \textbf{Mesh connectivity}: Static face connectivity stored per object for visualization only.
    \item \textbf{FPS anchors}: Deterministic Farthest Point Sampling anchors; the reported model uses $N_a{=}4$ anchors per object.
\end{itemize}

Physics parameters are consistent across all configurations: cube mass $m{=}1.0$\,kg, friction coefficient $\mu{=}0.5$, and restitution $\epsilon{=}0.3$. Each cube mesh contains 8 vertices (corner points), yielding total vertex counts of 224 ($28{\times}8$), 520 ($65{\times}8$), 1008 ($126{\times}8$), and 1736 ($217{\times}8$) for the four configurations, respectively.

\subsection{Training Configuration}
\label{supp_sec:wb_training}

We train \name on all four WreckingBall configurations jointly, using the mixed dataset of 72 training samples. The model uses the same object-anchor architecture as in the main MOVi experiments, including FiLM conditioning on $[s,s^2]$. It uses $N_a{=}4$ FPS anchors per object, and the maximum number of objects per frame is set to 220 to accommodate the largest configuration (217 objects) with a buffer.
Training uses a step size $s{=}10$ for temporal prediction, with a sequence length of 8 frames per training sample. The batch size is set to 3 due to the large object count. We do not use curriculum learning in this experiment.

The object-level tokenization enables efficient scaling: attention complexity grows as $O(M^2)$ in object count rather than $O((MN_v)^2)$ in vertex count. For WreckingBall-217 with 1736 total vertices, vertex-level attention would require ${\sim}3$M attention pairs, while our object-level design requires only ${\sim}54$K pairs (217 objects $+$ 16 registers), a $55{\times}$ reduction.

\section{Controllable Articulated Body Simulation}
\label{supp_sec:articulated}

To provide a preliminary test of \name beyond independent rigid objects, we construct two articulated-body datasets with high-level control inputs. We treat each body part as an object-level component and condition the model on command signals.

\subsection{Dataset Generation}

\paragraph{ASE HumanoidHeading.}
We generate training data from the Adversarial Skill Embeddings (ASE)~\cite{peng2022ase} framework using the HumanoidHeading task. The humanoid character is equipped with a sword and shield, comprising $M{=}17$ articulated body parts (15 body links plus the sword and shield). Data generation employs a hierarchical reinforcement learning architecture with pre-trained checkpoints: a Low-Level Controller (LLC) that generates joint torques $\boldsymbol{\tau}\!\in\!\mathbb{R}^{N_{\text{dof}}}$ from latent skill codes, and a High-Level Controller (HLC) that outputs latent skill codes conditioned on heading objectives.

Each trajectory spans $T{=}300$ frames at 30\,Hz (10 seconds of simulation). Per-frame control signals $\mathbf{c}_t\!\in\!\mathbb{R}^{5}$ include:
\begin{itemize}
    \item Target speed $v_{\text{tar}}\!\in\!\mathbb{R}$ (locomotion velocity magnitude)
    \item Target direction $\mathbf{d}_{\text{tar}}\!\in\!\mathbb{R}^{2}$ (movement direction unit vector)
    \item Target facing direction $\mathbf{f}_{\text{tar}}\!\in\!\mathbb{R}^{2}$ (heading orientation unit vector)
\end{itemize}
Meshes are simplified to a maximum of 300 vertices per body part using quadric edge collapse decimation. Mesh connectivity is used only for visualization; the model consumes point positions and per-point features, without surface normals.

\paragraph{G1 Steering.}
We generate data from the MimicKit framework using the Unitree G1 Steering task. The G1 robot comprises $M{=}31$ articulated body parts with significantly more complex geometry than ASE humanoids. Data generation uses the Adversarial Motion Priors (AMP)~\cite{peng2021amp} steering policy trained for directional locomotion.

Each trajectory spans $T{=}300$ frames at 30\,Hz. Per-frame control signals $\mathbf{c}_t\!\in\!\mathbb{R}^{3}$ include:
\begin{itemize}
    \item Target speed $v_{\text{tar}}\!\in\!\mathbb{R}$ (locomotion velocity magnitude)
    \item Target direction $\mathbf{d}_{\text{tar}}\!\in\!\mathbb{R}^{2}$ (movement direction unit vector)
\end{itemize}
For G1, each body-part object is represented by 200 points sampled from the original robot mesh, without surface normals. Joint positions $\mathbf{q}\!\in\!\mathbb{R}^{N_{\text{dof}}}$ are stored for kinematic analysis.

\begin{table}[h]
  \centering
  \caption{\textbf{Articulated Body Dataset Statistics.}}
  \label{tab:humanoid_datasets}
  \vspace{1mm}
  \small
  \begin{tabular}{lcc}
  \toprule
  \textbf{Property} & \textbf{ASE Heading} & \textbf{G1 Steering} \\
  \midrule
  Source Framework & ASE~\cite{peng2022ase} & MimicKit (AMP) \\
  Character & Humanoid + Sword/Shield & Unitree G1 \\
  Body Parts ($M$) & 17 & 31 \\
  Max Vertices/Part & 300 & 200 \\
  Control Dim & 5 & 3 \\
  Frames ($T$) & 300 & 300 \\
  Frequency & 30\,Hz & 30\,Hz \\
  \bottomrule
  \end{tabular}
\end{table}

\subsection{Implementation Details}

\paragraph{Control Signal Integration.}
To condition \name on high-level control signals, we use the same FiLM mechanism as temporal step-size conditioning. The per-vertex encoder input remains 12D; the FiLM conditioning vector is augmented from $[s,s^2]$ to $[s,s^2,\mathbf{c}_t]$, where $\mathbf{c}_t\!\in\!\mathbb{R}^{5}$ for ASE and $\mathbf{c}_t\!\in\!\mathbb{R}^{3}$ for G1. Thus, control commands modulate the encoder/decoder features through FiLM rather than being concatenated to every point feature.

\paragraph{Mesh Processing Pipeline.}
Raw data export proceeds in two stages:
\begin{enumerate}
    \item \textbf{Physics Export:} We run the trained policy in Isaac Gym~\cite{makoviychuk2021isaac} and record per-frame rigid-body transforms (position and quaternion) for each body part.
    \item \textbf{Mesh Conversion:} Body part transforms are applied to reference meshes from MuJoCo XML files. Meshes exceeding vertex limits undergo quadric decimation.
\end{enumerate}

\paragraph{Temporal Subsampling.}
We train with step size $s{=}10$ for both ASE and G1 (3\,Hz effective rate) to capture meaningful locomotion dynamics rather than high-frequency contact oscillations.

\paragraph{Training Configuration.}
We use $N_a{=}4$ FPS anchors per body part and the same object-anchor architecture as in the main rigid-body experiments, including FiLM conditioning. The only additional conditions are the high-level control signals injected through FiLM. The batch size is 4 for both datasets. We present this result as an extension study rather than the primary evidence for the method.

\section{Additional Ablations and Analyses}
\label{supp_sec:additional}

\subsection{Physical Parameters}
\label{supp_sec:phys_ablation}
\begin{table}[!htb]
  \centering
  \caption{\textbf{Physical parameters of the objects} (all scenes of each dataset). Friction and restitution follow the material: 0.4 and 0.3 for metal, 0.8 and 0.7 for rubber, and 94\% of the scenes of MOVi-A, MOVi-B and MOVi-Sphere hold both materials. The mass depends on the shape, the size and the material; in MOVi-C, it is the volume of the unscaled asset.}
  \label{tab:phys_stats}
  \vspace{1mm}
  \small
  \begin{tabular}{lccccc}
  \toprule
  \textbf{Dataset} & \textbf{Friction} & \textbf{Restitution} & \textbf{Mass} & \textbf{Distinct masses} & \textbf{Metal objects} \\
  \midrule
  MOVi-A      & 0.4 / 0.8 & 0.3 / 0.7 & 0.197--7.36 & 12 & 50.1\% \\
  MOVi-B      & 0.4 / 0.8 & 0.3 / 0.7 & 0.099--7.35 & 7{,}764 (all) & 49.5\% \\
  MOVi-Sphere & 0.4 / 0.8 & 0.3 / 0.7 & 0.197--3.87 & 4 & 50.8\% \\
  MOVi-C      & 0.5 & 0.5 & $1.0{\times}10^{-5}$--0.036 & 517 of 768 & -- \\
  \bottomrule
  \end{tabular}
\end{table}
\name\ reads the mass, friction and restitution of each object (Sec.~\ref{sec:method}); Tab.~\ref{tab:phys_stats} lists how they vary between objects. To measure how much it relies on them, we train on MOVi-B a model that gives every object the training-set means of the three parameters ($m{=}1.12$, $\mu{=}0.60$, $\epsilon{=}0.50$), with the architecture and training recipe of our model, and compare the two on the test scenes (Tab.~\ref{tab:phys_ablation}: w/o and w/ physical parameter modeling). Without the physical parameters, the position error is 8 to 25\% higher at step size 1 over frames 25 to 400 ($1.19\times$ at 100 frames, 95\% interval of a paired bootstrap over scenes $[1.10, 1.31]$) and 8 to 31\% higher at step sizes 5 and 10, while the orientation error changes by at most 7\%: \name\ uses the parameters, mainly to place the objects.

\begin{table}[!htb]
  \centering
  \caption{\textbf{Physical parameters} (MOVi-B test scenes): our model (w/ physical parameter modeling) and the model trained with the training-set means of the parameters for every object (w/o). Position RMSE (m)\,/\,orientation RMSE ($^{\circ}$) at 50, 75 and 100 frames; step size 10 predicts no 75-frame entry here.}
  \label{tab:phys_ablation}
  \vspace{1mm}
  \small
  \begin{tabular}{llccc}
  \toprule
  \textbf{Step} & \textbf{Model} & \textbf{50} & \textbf{75} & \textbf{100} \\
  \midrule
  1 & w/ physical parameter modeling & 0.030\,/\,3.24 & 0.061\,/\,7.14 & 0.108\,/\,12.65 \\
  1 & w/o physical parameter modeling & 0.033\,/\,3.35 & 0.071\,/\,7.39 & 0.129\,/\,13.14 \\
  \midrule
  5 & w/ physical parameter modeling & 0.020\,/\,2.70 & 0.047\,/\,6.31 & 0.090\,/\,11.57 \\
  5 & w/o physical parameter modeling & 0.024\,/\,2.75 & 0.057\,/\,6.43 & 0.110\,/\,11.89 \\
  \midrule
  10 & w/ physical parameter modeling & 0.012\,/\,1.23 & -- & 0.071\,/\,8.82 \\
  10 & w/o physical parameter modeling & 0.014\,/\,1.32 & -- & 0.088\,/\,9.10 \\
  \bottomrule
  \end{tabular}
\end{table}

\subsection{Encoder}
\label{supp_sec:encoder_ablation}
We replace the hierarchical PointNet encoder (Sec.~\ref{sec:method}) with a set-abstraction encoder in the style of PointNet++~\cite{qi2017pointnet++} (two set-abstraction levels with farthest point sampling and $k$-nearest-neighbor grouping, followed by feature propagation to all points) and train it on MOVi-B with the training recipe of our model. Tab.~\ref{tab:enc_ablation} compares it with our model on the test scenes. The set-abstraction encoder has a position error 1 to 13\% higher at step size 1 over frames 25 to 400 ($1.06\times$ at 100 frames, 95\% interval of a paired bootstrap over scenes $[0.97, 1.15]$), up to 22\% higher at step size 5 and 3 to 9\% higher at step size 10; its orientation error ranges from 1\% lower to 18\% higher. The set-abstraction encoder thus does not improve on the hierarchical encoder of \name.

\begin{table}[!htb]
  \centering
  \caption{\textbf{Encoder} (MOVi-B test scenes): our model and the model with a set-abstraction encoder. Position RMSE (m)\,/\,orientation RMSE ($^{\circ}$) at 50, 75 and 100 frames; step size 10 predicts no 75-frame entry here.}
  \label{tab:enc_ablation}
  \vspace{1mm}
  \small
  \begin{tabular}{llccc}
  \toprule
  \textbf{Step} & \textbf{Model} & \textbf{50} & \textbf{75} & \textbf{100} \\
  \midrule
  1 & Our model & 0.030\,/\,3.24 & 0.061\,/\,7.14 & 0.108\,/\,12.65 \\
  1 & Set-abstraction encoder & 0.032\,/\,3.76 & 0.065\,/\,7.91 & 0.114\,/\,13.66 \\
  \midrule
  5 & Our model & 0.020\,/\,2.70 & 0.047\,/\,6.31 & 0.090\,/\,11.57 \\
  5 & Set-abstraction encoder & 0.022\,/\,2.83 & 0.049\,/\,6.34 & 0.092\,/\,11.51 \\
  \midrule
  10 & Our model & 0.012\,/\,1.23 & -- & 0.071\,/\,8.82 \\
  10 & Set-abstraction encoder & 0.012\,/\,1.28 & -- & 0.077\,/\,8.74 \\
  \bottomrule
  \end{tabular}
\end{table}

\FloatBarrier
\subsection{Attention and Interaction Graphs}
\label{supp_sec:attention}
\begin{figure}[!p]
  \centering
  \includegraphics[width=\linewidth]{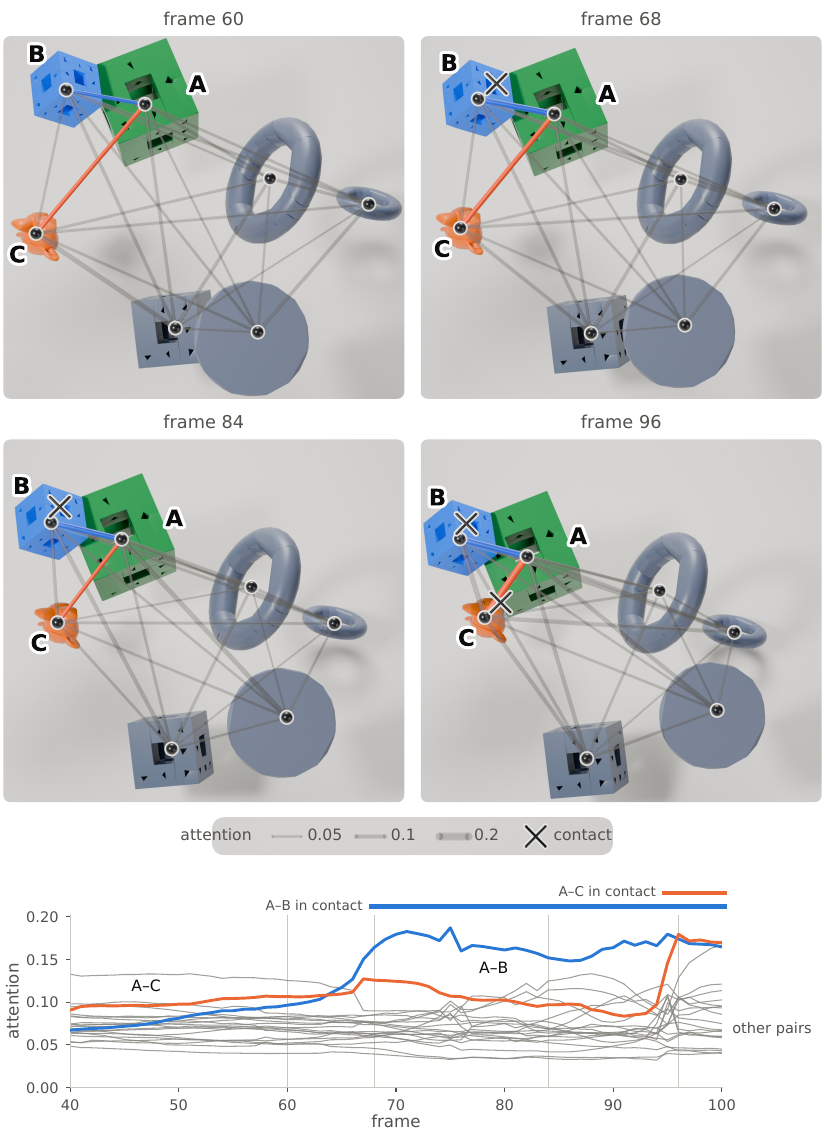}
  \caption{\textbf{Interaction graph of a MOVi-B test scene.} Top: the predicted states at four frames of the rollout (step size 1), rendered from one elevated view (objects A, B and C in green, blue and orange); each edge joins the anchor centroids of two objects and its width increases linearly with the attention between them, averaged over heads, layers and both directions; crosses mark where two objects touch. Bottom: attention between object A and objects B and C, and between all other pairs (gray), over frames 40 to 100; bars mark the frames at which the two objects are in contact in the predicted state. In the ground truth, A and B touch at frames 68--73 and A and C at frames 94--96; in the rollout, A and B remain in contact (overlapping) after frame 73.}
  \label{supp_fig:attention}
\end{figure}
We relate the object-level self-attention of the decoder to contact. We roll out the models of Tab.~\ref{tab:main_comparison} on the 120 test scenes of each dataset (step size 1) and average the attention between two object tokens over heads and both directions, at frames up to 100. Two objects are in contact when their meshes are closer than 1\,cm, and never close when they stay farther than 10\,cm within $\pm5$ frames.

In every decoder layer and on all three datasets, objects in contact receive more attention than objects that are never close (over the four layers $1.96\times$ on MOVi-B, $1.78\times$ on MOVi-Sphere and $1.28\times$ on MOVi-A; every 95\% interval over scenes lies above 1), also at the same distance between their anchor centroids ($1.59\times$, $1.75\times$ and $1.22\times$). On MOVi-B, the attention grows as two objects approach (0.071 beyond 20\,cm, 0.140 below 5\,mm), and in the scene of Fig.~\ref{supp_fig:attention} it rises within a few frames of each contact. These are correlations, and over all 476 rollout steps the ratio is lower (1.12 to 1.24).

\section{Discussion on Temporal Step Sizes}
\label{supp_sec:step_size_discussion}

Step-size conditioning controls the effective temporal resolution of the learned simulator. A small step size is closest to standard one-step rollout protocols and returns dense future states, but a fixed physical horizon then requires many autoregressive calls, increasing both computation and accumulated rollout error. A larger step advances the learned world model over a longer interval per call, allowing it to expose sparse long-horizon future states more efficiently.

This trade-off is useful for planning. In many robotics settings, such as model-predictive planning or trajectory optimization, the planner needs to compare possible future object configurations and interaction outcomes, but does not always require every high-frequency contact frame. Coarse large-step rollouts can therefore provide a cheap long-horizon search signal, while smaller steps remain useful for local control, contact refinement, and short-horizon correction. A natural extension is adaptive time stepping: use large steps to explore candidate futures, then refine selected plans with smaller steps or uncertainty-aware rollouts.

\section{Limitations, Future Work, and Impact}
\label{supp_sec:limitations_future}

\subsection{Limitations and Future Work}
\name is designed for object-level rigid-body dynamics from point clouds and assumes object labels that indicate which points belong to each object. Although we include partial point-cloud results, by masking points inside each object's bounding box (Appendix~\ref{supp_sec:more_vis}) and from a single depth camera with completion (Appendix~\ref{supp_sec:partial_depth}), severe partial observations can become challenging when the visible points capture too little of the object's shape. Future work could study stronger occlusions, real sensor noise, online object segmentation from raw observations, mixed rigid--deformable scenes, and more fine-grained adaptive time stepping.

The rigid projection step exactly preserves intra-object distances, but contact handling is still learned from data rather than solved with an explicit complementarity-based physics engine. This design keeps the model mesh-free and efficient, while leaving room for contact-aware losses, lightweight correction layers, or hybrid learned-analytic constraints in scenes with unusually sharp contact regimes. Our current study also focuses primarily on rigid objects, with articulated bodies treated as collections of object-level parts; extending the same representation to mixed rigid--deformable scenes is a natural next step.

\name also takes the mass, friction and restitution of each object as inputs, and all reported results use the values of the simulator; Appendix~\ref{supp_sec:phys_ablation} measures the effect of withholding them. In practice these quantities would have to be estimated, for example by online system identification from the observed motion, which fits naturally into the object-centric state. The per-point velocities are differences between consecutive frames, so the inputs assume temporally consistent point samples of each object.

The errors grow with the rollout horizon (Tab.~\ref{tab:ablation_steps}), and we give no bound on this growth; larger step sizes reduce it but do not predict the frames in between. Dense, sustained contact remains the most difficult regime. The runtimes we report are those of the model on one GPU; a deployed system adds the cost of perception, segmentation and point completion.

\subsection{Impact Statement}
This work aims to improve machine learning methods for physical dynamics modeling, especially mesh-free simulation of rigid-body contact from point clouds. Efficient learned simulation can benefit robotics, graphics, and embodied AI by reducing the cost of rollout prediction, supporting planning over object interactions, and enabling controllable physical reasoning in virtual environments.

The main deployment consideration is reliability under distribution shift. Models derived from this work should be validated under the intended geometry, material, and contact conditions before being used in safety-relevant control loops. In high-stakes settings, predictions should be paired with uncertainty checks, safety constraints, and human or system-level oversight. The method operates on geometric state data rather than personal data, so we do not anticipate direct privacy risks from the core technique.